# In Search of Excellence: SHOA as a Competitive Shrike Optimization Algorithm for Multimodal Problems

**Hanan K. AbdulKarim[1], Tarik A. Rashid[2], Member, IEEE**

[1]Software Engineering Department, Engineering College, University of Salahaddin - Erbil, Erbil, Iraq
[2]Computer Science and Engineering Department, University of Kurdistan Hewler, Erbil, Iraq

Corresponding author: Hanan K. AbdulKarim (email: hanan.abdulkarim@su.edu.krd)

**ABSTRACT** In this paper, a swarm intelligence optimization algorithm is proposed as the Shrike Optimization Algorithm (SHOA). Many creatures living in a group and surviving for the next generation randomly search for food; they follow the best one in the swarm, called swarm intelligence. Swarm-based algorithms are designed to mimic creatures' behaviors, but in the multi-modal problem competition, they cannot find optimal solutions in some difficult cases. The main inspiration for the proposed algorithm is taken from the swarming behaviors of shrike birds in nature. The shrike birds are migrating from their territory to survive. However, the SHOA mimics the surviving behavior of shrike birds for living, adaptation, and breeding. Two parts of optimization exploration and exploitation are designed by modelling shrike breeding and searching for foods to feed nestlings until they get ready to fly and live independently. This paper is a mathematical model for the SHOA to perform optimization. The SHOA benchmarked 19 well-known mathematical test functions, 10 from CEC-2019, and 12 from CEC-2022 most recent test functions, a total of 41 competitive mathematical test functions benchmarked and four real-world engineering problems with different conditions, both constrained and unconstrained. The statistical results obtained from the Wilcoxon sum ranking and Fridman test show that SHOA has a significant statistical superiority in handling the test benchmarks compared to competitor algorithms in multi-modal problems. The results for engineering optimization problems show the SHOA outperforms other nature-inspired algorithms in many cases.

## I. INTRODUCTION

Optimization techniques have become important in the last few decades. Optimization is finding the best optimal or semi-optimal solution by achieving a specific objective without violating constraints. In some cases, no objective functions exist, but a feasible solution depending on constraints is an optimal solution, called a feasibility problem. Many complex and rough-solvable problems in engineering, science, medicine, statistics, and computer science have been solved by optimization algorithms within a short time. Mathematical calculation and programs have been used to solve such a problem, but recently, for solving complex problems, some meta-heuristic optimization algorithms have been used to find acceptable solutions. Many optimization algorithms are nature-inspired algorithms designed by mimicking creatures from nature; many of those algorithms depend on swarms' social behavior and are called swarm-based algorithms.

Optimization algorithms have been classified as single-based and population-based. Single-based optimization searches for an optimum solution using a single solution like simulated Annealing (SA), Hill Climbing (HC), Variable Neighborhood Search (VNS), and Tabu Search (TS) [1–4], while the population-based optimization algorithms use a group of solutions as a population and search around the number of the neighbors of the solutions in the search space, it should have good exploration and exploitation techniques to not trap in the local optima, population-based like Genetic Algorithm (GA) [5], Differential Evaluation (DE) [6], Genetic Programming (GP) [7], and swarm-based algorithms.

Swarm-based algorithms are stochastic because they work on the Swarm Intelligence (SI) of the creature's behavior. Ant Colony Optimization (ACO) [8] is an old algorithm that studies the collective behavior of ants searching for food sources. It simply translates the fact that every ant has its own decision for foraging on a specific path, each ant signs the path by pheromone when it transits to the food source, and it will add pheromone again when returning to the nest, so other ants will take a path with higher pheromone and leave their path, then the shortest path will be accomplished by leaving the low pheromone path and use the higher-level pheromones path. Ant System (AS) [9,10] applied to solve various combinatorial optimization problems. The application of AS includes the Traveling Salesman Problem (TSP), the Quadratic Assignment Problem (QAP), and the Job-shop Scheduling

Problem (JSP), it shows the ability to solve those problems, also applied in the classification field [11], and cloud computing [12]. Recently Ant Nesting Algorithm was proposed [13] searching to build nests and deposit foods.

Particle Swarm Optimization (PSO) [14] mimics the inspiration of SI of birds, and fish while the author considered birds, simple techniques were used that birds follow the flock fly direction, the best food source obtained so far, and the best food sources that the swarm found, simply it uses rules to find the best solution in the search space, and it is applied in many fields of design, image processing, and others [15,16] which successfully improves solutions. The Society and Civilization algorithm [17] is the adaption of societies simulated for optimization problem-solving. Artificial Bee Colony (ABC) works on honey bees finding food sources in [18,19], it works on how explored bees find food sources and share information with employed bees, the onlooker bees exploit food sources more to find better sources and keep the best food source, ABC outperforms many optimization algorithms for some optimization problems of global optimization, feature selection, neural network fields, vehicle routing [20–24]. Fitness Dependent Optimizer (FDO) also working on bees foraging behavior is proposed for optimization problems [25]. Bacterial Foraging behavior (BFO) the bacterial foraging behavior has been a source for development, applied for electrical power filter problems, and designed fuzzy control for the system [26–28]. In Firefly Algorithm (FA), the flashing light and attractiveness of fireflies were formulated as FA algorithm, used to solve multi-modal problems, design structure, and many other applications [29,30]. The moth-flame Optimization (MFO) Algorithm [31] was developed by studying moths' navigation in nature and how they move around lights. Solving problems with clustering suffers from exploration the MFO is added to handle the clustering problem [32]. Recently some new population-based metaheuristic algorithms have been proposed by researchers like new artificial protozoa optimizer (APO) inspired by biology [33], the Horned Lizard Optimization Algorithm (HLOA) used the defensive strategies of the horned lizard reptile[34] , the Black Winged Kite birds' skills in the fields of hunting and migrating are modeled as Black Winged Kite Algorithm (BKA) [35], the Hiking Optimization Algorithm (HOA), which hikers travel uphill, and HOA seeks to find the local or global optimal solution to optimization issues[36] In the exploration, the authors mention trigonometric functions that are used to search space globally to cover all space and avoid local optima. Although search space coverage uses sin and cosine to guarantee searching for all solutions, while in the exploitation, refinement and local search on founded solutions will be done to improve accuracy and convergence to the global optimum [37].

Nature-inspired algorithms have demonstrated exceptional performance in optimization problems, particularly multimodal problems. Soccer inspired metaheuristic-based sports concepts, they became popular because they developed inspiration and convert them to algorithm steps to solve problems. Recently developed algorithms are studied and methods and concepts [38]. The Squirrel Search Algorithm (SSA) is an optimizer that emulates the dynamic hunting behavior of southern flying squirrels [39] , the Marine Predator Algorithm (MPA) is a metaheuristic algorithm designed to emulate the hunting behavior exhibited by marine predators [40] . A new algorithm is proposed in [41] as a peacock algorithm that mimics the mating and hunting behaviors of peacock birds. The Cheetah (C) algorithm is inspired by the cheetah's foraging strategy [42]. Mountain Gazelle Optimizer (MGO) is an algorithm that takes inspiration from the social life of mountain gazelles [43]. Since the reviewed methods initially were proposed, the researchers have worked to enhance or implement them in many domains and for various challenges [44–52].

This paper proposes a swarm-based SHOA refers to a bird-inspired class, to increase the number of solved multi-modal and complex problems because none of the optimization algorithms can solve all problems. Depending on the nature of the problem, a specific algorithm must be applied to find the best solution. Intensification and diversification are the essential components of meta-heuristic algorithms. The main contributions of this study are:

1. The proposed SHOA is designed for multi-modal problems by finding many local optima and keeping them to produce global optima because multi-modal problems have many local optima and many optimization algorithms lack the ability to find the global optima.
2. In the SHOA mathematical proposal produced, depending on the shrike bird's physical simulations of a parent bird's dominance in a specific life stage, the roles of female and male birds were separated depending on reality and lifestyle.
3. The SHOA, applying randomization will diverge the algorithm from the current solution, which is considered a local optimum, and redirect the algorithm to search the space globally to increase diversity, while finding a solution during the local search by choosing the best solution so far will converge the algorithm to an optimal solution, increasing convergence.

The remainder of this article is structured as follows: (Part II) represents the literature viewed, and important points are discussed,  (Part III) presents inspiration from shrike birds and a mathematical model for the proposed SHOA (Part IV) results and discussion on comparative benchmarks and some competitive functions, and (Part V) SHOA applied real-world cases, studies as engineering problems and the performance compared with other optimization algorithms, finally (Part VI) conclude the work of this study and show direction for coming studies.

A. Literature Review

Metaheuristic optimization algorithms are nature-inspired algorithms, they study techniques and rules of the creature's classified, in this study some of the classes are prepared and noted to classify algorithms rather than categories mentioned by researchers. Nature-inspired algorithms are refined to classes with some examples shown in Figure 1. Some of the recently proposed algorithms with classes are:

1. Animal-inspired classes are sub-grouped into the bird, mammal, fish, insect, inspired, some examples are; Eurasian Oystercatcher Optimizer (EOO) [53], White Shark Optimizer (WSO) [54], Fox Optimizer (FOX) [55], Orca Optimization Algorithm (OOA)[56], Walrus Optimization Algorithm (WaOA) [57], Aphid-Ant Mutualism (AAM)[58], Fire Hawk Optimizer (FHO)[59], Honey Badger Algorithm (HBA) [60], Tunicate Swarm Algorithm (TSA)[61], Pufferfish Optimization Algorithm (POA) [62], Marine Predator Algorithm (MPA), Horned Lizard Optimization Algorithm (HLOA).
2. Plant, Microorganism, Physics, Human Activity, Mathematics, Algorithm-Specific and miscellaneous classes, all classes are shown in Figure 1, some examples of recently proposed algorithms are:
    a. Water Wheel Plant Algorithm (WWPA) [63]
    b. Artificial Protozoa Optimizer (APO)
    c. Black Hole Mechanics Optimization (BHMO)[64]
    d. Chef-Based Optimization Algorithm (CBOA)[65]
    e. Gradient-Based Optimizer (GBO)[66]
    f. One-to-One-Based Optimizer [67]
    g. PID-based Search Algorithm (PSA) [68]
    h. Ali Baba and the Forty Thieves Optimization (AFT) [69]

B. **MULTIMODAL OPTIMIZATION AND PARAMETER TUNING**

Multimodal optimization problems (MMOPs) necessitate the simultaneous search for several optimum solutions. Assert that the algorithm needs to broaden its population diversity to identify more global optima, and enhance its refinement capabilities to boost the accuracy of the discovered solutions [70]. The group of multimodal approaches using the methods of speciation and crowding-niching. Whereas speciation separates the population into individuals of the same species, crowding-niching splits the population into niches occupied by various species. Several multimodal techniques, including the Fitness Sharing (FS) technique, consider sharing models and similarity functions [71].

The researchers have implemented several niching strategies to divide the population into distinct niches, each responsible for conducting searches on one or more peaks [70]. Real-world design challenges known as constrained numerical optimization problems (CNOPs) require a feasible, ultimate optimized solution, and restrictions act as roadblocks to potential solutions. When it comes to addressing conventional unconstrained numerical optimization problems (UNOPs), nature-inspired meta-heuristics are popular. Effective algorithms require population diversity to accomplish design space exploration, but they reduce diversity through optimization to leverage the space's global optimum [72]. Handling the niche centre distinction (NCD) problem as an optimization problem. Performance measures assess success, accuracy, feasibility[72].Many contexts, including data mining, power systems, pattern recognition, and vehicle routing issues, have used MMO algorithms. Several researchers have proposed multi-objective evolutionary optimization strategies for solving MMOPs using bi-objective problems. In addition to the steps used in Evolutionary Algorithms (EAs), MMO algorithms employ additional strategies to converge on numerous solutions. Authors suggested EAs use one of two prevalent niching techniques: species-based DE (SDE) or crowding-based DE (CDE) [70]. Researchers have used a stable mutation approach to create new individuals, the SoftMax function to determine individual probabilities, and an archive technique to retain stagnant individuals [73].

In [74], the authors solve MMOPs using Distributed Individuals for Multiple Peaks (DIMP) used with DE, by applying age to each individual, DIMP allows every individual to function as a dispersed unit to monitor a peak, avoiding the challenges associated with population division and preserve enough variation to find new peaks throughout their lifetime.

Differential evolution (DE) is an efficient yet straightforward approach extensively researched for both MMOPs and single optimum optimization problems [74].

The authors in [75], applied the two-phase stream clustering algorithm based on fitness proportionate sharing to produce data for MMOP. The authors then developed a novel dynamic clustering algorithm to extract the cluster structure automatically from scratch and approximate the density distribution of the data stream using a recursive lower bound of the Gaussian kernel function. Applying the Cluster-Chaotic-Optimization (CCO) approach for a specific optimization issue, extending its capabilities to discover, register, and retain many optima effectively [71].

Because parameter tuning is a hyper-optimization problem, it is particularly difficult when tweaking optimization techniques. Currently, it is unclear how to tune parameters effectively and how to control parameters properly for any given algorithm, and a given set of problems. Although a good tuning tool apply, the tuned parameters may not perform well for other problems, or for different types with unknown optimality [76]. Parameter tuning may be done in a variety of ways, such as hyper-optimization, which applies the ideal parameter setting n* after the algorithm has been tuned for the given issue M. A parallel or loop structure for tuning and problem-solving will be used repeatedly as an additional method for parameter tuning [76]. The approach known as self-adaptive multi-population (SAMP) involves the dynamic addition and delete of populations according to their variety. The free population, the initial population in this approach, is a single randomly initiated population. After evolution, the authors declared a solution to have converged if the gap between them decreases below a specific threshold.

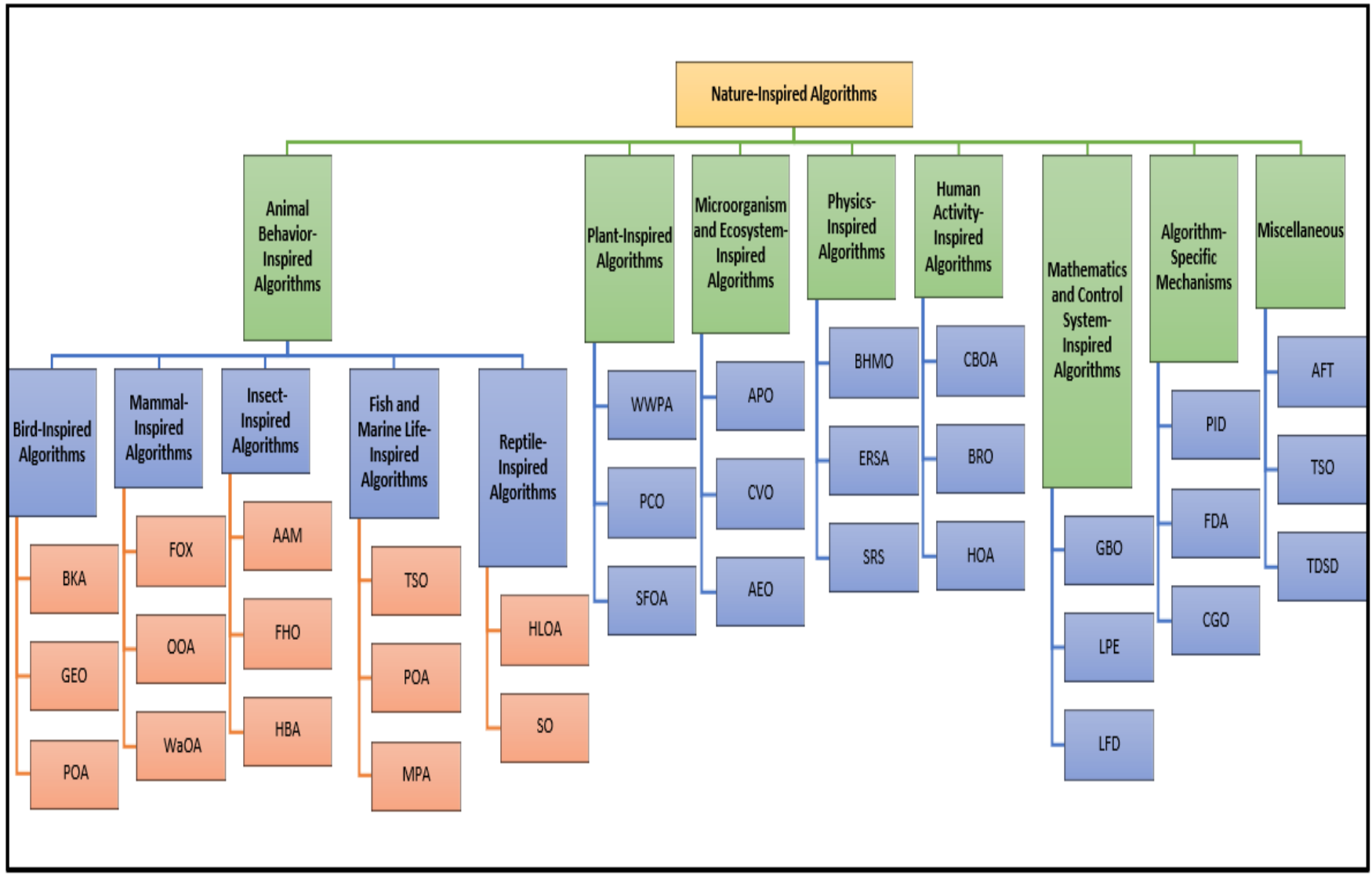

*Figure 1 Classification of metaheuristic optimization Algorithm*

If all current populations have converged, new populations will be randomly introduced. SAMP maintains at least one free population at all times to prevent the algorithm from becoming stuck in local optima [77]. The researchers also researched random partitioning, a potential population partitioning approach that divides a single population into several smaller sub-populations at random using seed-based partitioning with fixed seeds and random partitioning with a master population [77]. The significance of initialization is crucial for the accuracy and rate of convergence of certain algorithms. Researchers should utilize various initialization techniques for different situations, as starting solutions may impact the effectiveness of the optimization algorithm. Uniform distributions are not the optimal initialization strategy for all functions [78]. Robust Optimization Over Time (ROOT) is a discipline that focuses on investigating and advancing algorithms, incorporating the principles of both adaptive and robust optimization [79]. The No Free Lunch (NFL) theorem states that there are no metaheuristic algorithms or optimization strategies available to handle the issue optimally. While metaheuristic optimization approaches can be helpful in solving some issues, they can also be ineffective in solving other difficulties. There is still room for improvement in the field of metaheuristic optimization algorithms, and several academics are working to build new metaheuristic algorithms [80]. Applying diversity measures improves the understanding and efficiency of algorithms. Researchers suggest replacement and exclusion operator strategies. By randomly reinitializing the population with a less-fit solution, the inclusion operator in multi-population swarm algorithms preserves population variety. There is a replacement operator that creates new, and randomly generated solutions to replace the ones that already exist. The majority of diversity-increasing methods help an algorithm's fundamental structure to be modified [81]. Researchers have found the use of the enhanced search method with various swarm algorithms that use Cauchy, Levy, and uniform distributions [82].

The effective application of swarm-based algorithms by the scientific and business communities has demonstrated the worth of these methods in practice. The benefits of SI-based algorithms are the reasoning success of the algorithms mentioned before. Swarm-based optimization methods work with groups as a population and have some randomness during searching for a solution. Despite all optimization algorithms, there is no universal algorithm used to solve all optimization problems. Still, some algorithms outperform others in many types of optimization problems. The researchers are working to find an algorithm that outperforms other algorithms for most of the problems or find new algorithms that can solve unsolved problems.

## II. SHRIKE BIRDS

### A. SOURCES OF INSPIRATION FOR SHRIKES

The Laniidae family of passerine birds includes shrikes, which are distinguished by their propensity to impale their flesh on thorns after capturing insects, small birds, or animals. The shrikes are two genera with 34 species distributed throughout the world. In North America, there is a member of the Shrike family called Loggerhead Shrikes. Loggerhead shrikes, also called butcherbirds and migrating shrikes, reach a weight of roughly 48 grams [83–85]. Within the Laniidae family, this remarkable bird is rather huge, and its large head may have contributed to its unique name. Males and females have similar appearances; it is difficult to distinguish between them. They have black, white, and grey markings on their bodies and a black mask that covers their eyes [86].

Over its range, the loggerhead's appearance varies slightly by region. Loggerheads eat mainly small vertebrates and small mammals. They live, migrate, eat in population, and use cooperative breeding [84]. Make nests on the trees; the female will deposit between four and seven eggs in a clutch, which she will then incubate for roughly sixteen days [86]. For a period of seventeen to twenty days, both parents are responsible for taking care of the nestlings. After leaving the nest, the young birds remain close to their parents for three weeks, during which time they get food from both parents, develop their flight, and at night return to be warmed by the parents. For more information, return to reference [83]. The population of the shrike bird life cycle is simulated in the Figure 2. There are three nests: A, P, and Q are the population of birds' nests; the nest AA parent will brood eggs at the nest (AB); the nestling will grow up and become adults ready to fly and later depend on themselves, the breeding and surviving of the birds of nest A shown from (A to C).

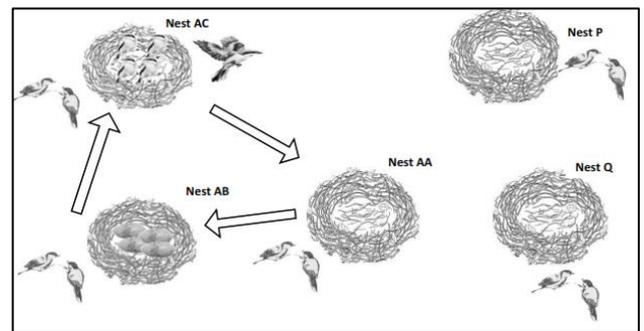

Figure 2 Shrike bird life cycle

### B. SHRIKE OPTIMIZATION ALGORITHM

Depending on the nesting and reproductive behavior of the shrike birds explained in the previous section, the shrikes live in a population out of the urban area; the population has many nests, and each nest starts with two birds as parents. The breeding and surviving behaviors of the shrikes were modeled by the Shrike Optimization Algorithm (SHOA). In Figures (3 and 4) the pseudo-code SHOA is presented. The pseudo-code clearly and simply describes the SHOA's execution process.

The SHOA start by initializing parameters, where N is the size of nests in the population, B is the number of eggs considered nestlings in each nest, and α is constant considered a natural factor affecting the bird. The search space of SHOA starts with a population of N nests, where each nest starts with two parent birds generated randomly.

```
Initialize parameters N , Max_Iteration , B, α , k
Initialize population nest_i = (i=1,2,..., N)
for each nest_i
    Generate P_j ( j=1,2) using equation (2)
while (! Max_Iteration)
    for each nest_i
        // every k generation check
        if (nest_i has only 2 birds)
            generate egg_j ( j=1,2,...,B) using Algorithm Fig(4)
        else
            choose best 2 birds and remove others
            generate egg_j ( j=1,2,...,B) using Algorithm Fig(4)
        get M_parent , F_parent ( nest_i )
        for each bird_j in nest_i feed
            generate random r using equation (5)
            if (bird_j is Parent )
                calculate Δfood_j using equation (6)
            else // phase 1 exploitation
                calculate Δfood_j using equation (7)
                calculate bird_j^(t+1) using equation (9)
                if( Fit_j^(t+1) better than Fit_j^(t) ) keep bird_j^(t+1)
                else
                    calculate Δfood_j using equation (8)
                    calculate bird_j^(t+1) using equation (9)
                    if( Fit_j^(t+1) better than Fit_j^(t) ) keep bird_j^(t+1)
                    else // phase 2 exploration
                        explore solution using equation (10)
        keep local best of nest_i
    end
    keep global best from all nest
endwhile
```

Figure 3 Pseudo-code SHOA Algorithm

```
Initialize F_Parent , M_Parent , B = nestling size
For each nestling_j
    generate random r ∈ [-1,1]
    calculate Δfood_j using equation (3)
    calculate nestling_j using equation (4)
    calculate Fit_j
    add to nest
```

Figure 4 Pseudo-code generates nestling steps

After the population is generated and nests are ready, and B number of nestlings will be generated. Population is represented as equation (1).

$$Population(N) = \begin{pmatrix} \begin{bmatrix} p_{im} & p_{if} \\ n_{ij} & n_{ij} \end{bmatrix} & \cdots & \begin{bmatrix} p_{im} & p_{if} \\ n_{ij} & n_{ij} \end{bmatrix} \\ \vdots & \ddots & \vdots \\ \begin{bmatrix} p_{im} & p_{if} \\ n_{ij} & n_{ij} \end{bmatrix} & \cdots & \begin{bmatrix} p_{im} & p_{if} \\ n_{ij} & n_{ij} \end{bmatrix} \end{pmatrix} \quad (1)$$

Where a population like a pool has N nests, each element in the population represents a nest, each nest_i has many solutions parent and nestling considered as a solution of the algorithm, where i = (1,2, ... N), and parents are randomly generated using equation (2).

$$p_i = LB + rand\ (UB - LB) \quad (2)$$

In the initialization process, after two birds are generated as parents for each nest, the fittest will be selected as dominance $M_{parent}$ is male parent, and other remains $F_{parent}$ as a female parent. In the breading phase, every nest generates B nestling using equations (3) and (4). Where $\Delta egg_j$ generate from both parents, and r is a random value in [-1,1], then $\Delta egg_j$ used to generate nestling_j , where i= (1,2, … B).

$$\Delta egg_j = (F_{parent} - M_{parent}) + r \quad (3)$$

$$nestling_j = F_{parent} + \Delta egg_j \quad (4)$$

The nestlings will depend on their parents; the male parent is dominant, which feeds the nestling, and the female, but the male feeds by itself only; the female also feeds by itself, and will feed nestling if they don't get food from the male parent. The idea of feeding nestling by a dominant parent leads the search to exploit solutions and converge to the optimum solution. Each nest has only two dominant solutions as a parent, they consider the first optimum and second optimum solution for the current nest, and each nestling gets feeds from the parent this is the solution exploit phase. In SHOA, after initialization nests and parameter parents should be specified depending on their objective function, and then r will be generated for each dimension using equation (5). The r parameter represents a natural factor in feeding, and the reason for the calculation of such a factor is to increase exploration.

$$r = e^{-2xt/T_{max}} \quad (5)$$

Where x is the dimension variable for bird_j, t is the current iteration, and $T_{max}$ is the maximum iteration allowed for running SHOA. Then using equation (6) each parent bird will feed itself.

$$\Delta food_j = bird_j * r \quad (6)$$

But for feeding nestlings, the Δfood is generated using formula (7), which $bird_j$ is the current bird state with a $M_{parent}$ is male parent bringing food.

$$\Delta food_j = r * (bird_j - M_{parent}) + bird_j \quad (7)$$

Whereas the nestlings didn't survive by food from the male parent, then they tried to survive through the female parent using formula (8), the same as formula (7), but r ∈ [-1,1], and sin(α), where α is used as a constant factor.

$$\Delta food_j = r * (bird_j - F_{parent}) + sin(\alpha) \quad (8)$$

After generating food, the birds' next status will be calculated using formula (9), which is the current state of birds getting food.

$$bird_j^{t+1} = bird_j^t + \Delta food_j \quad (9)$$

Calculate the fitness for each bird is better than the current state, then the current bird $bird_j^t$, will be updated with the new state as $bird_j^{t+1}$, the fittest one will survive for the next generation, not all birds get food at the same time. If any bird$_j$ does not get food from its parent it will survive using equation (10) to generate $\Delta food_j$, where r is randomly generated between [-1,1] and another variable parameter α = rand [0, dimension], α used as random variable to increase randomization, the sine of the variable will change over time depending on different values, this step is exploring the space by finding new solutions far from the current state, and randomly searching other possibilities, in this phase, the current solution will diverge from local best to generate new solution far away from parent.

$$bird_j^{t+1} = bird_j^t + (r * bird_j + sin(\alpha)) \quad (10)$$

The SHO algorithm keeps the best of each nest as the local best, then the population keeps the best from all local best as global. The idea of multi-modality can be solved using a group of solutions; where each nest has many birds, each k iteration the nest will regenerate after the old nestling finishes their nestling period time. The algorithm will keep just the two best birds as parents and remove all other birds as they die or fly far away from nests as they get ready to live independently. Every parent will generate a new nestling again and a new generation will update the current nest solution, the same algorithm execution process will continue till the stop condition.

Generation after generation of searches conducted using a randomly generated population. The Flow chart of SHOA specifies the process and SHOA's follow steps are presented in Figures (5,6).

### C. Time and Space Complexity

The computational time complexity of the SHOA encompasses the time and space complexity is taken into account. The time complexity of SHOA is affected by the initialization process and population updating as follows: The algorithm initialization process requires O(N × B) time, where, as mentioned, N is the number of nest members in the population and each nest has B birds.

Updating and calculating each nested element as a solution of the algorithms requires M iterations to complete algorithm evaluation O ((N× B )× M), where M is the maximum iteration of an algorithm. For every dimension, the updating of members requires an O ((N× B )× M × d) time, where d is the dimension, for every k iteration SHOA had to choose two best to regenerate nestling, the time considered as O (N log N). Overall time complexity:

O ( N×B ( 1 + M ( 1 + d ))) + O ( N log N).

The space complexity of SHO depends mainly on the population size (N), the number of solutions in each nest (B), and the number of dimensions to be solved (d).

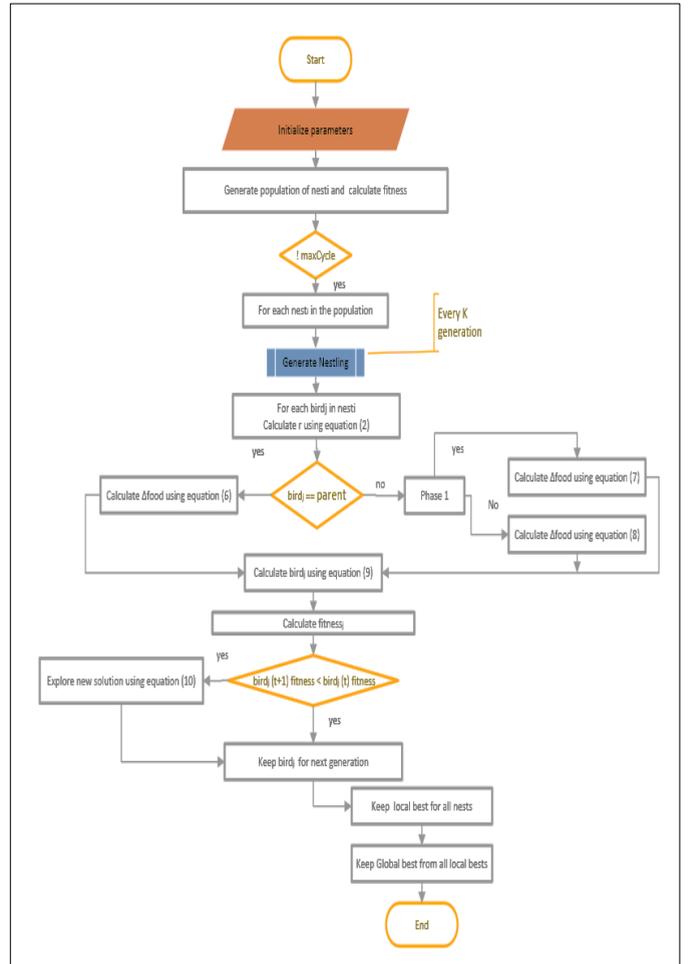

Figure 5 Flow chart of SHOA

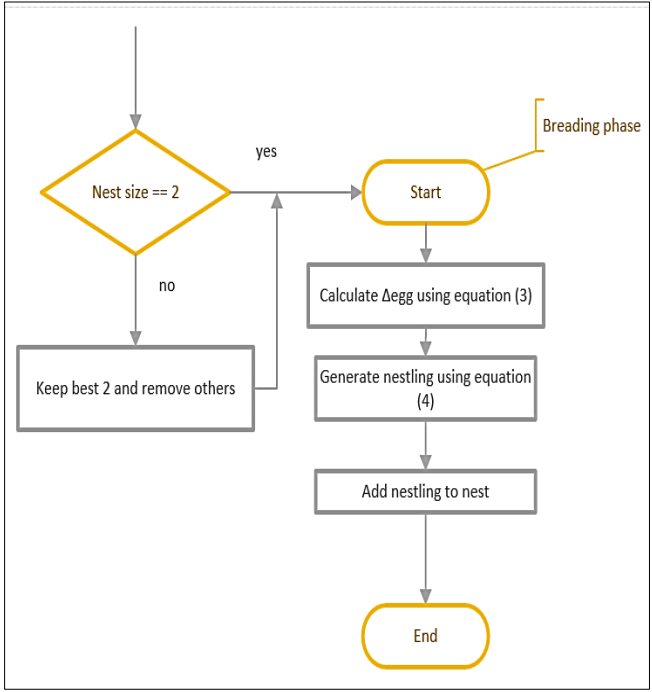

Figure 6 Flow chart of Generate Nestling

## III. IMPLEMENTATION AND RESULTS

number of global optimization test functions to show the performance of SHOA and the results compared with some well-developed optimization algorithms studied in the literature. Three groups of test functions are selected as uni-modal, multi-modal (simple, complex), 100-digit Challenge test functions [87–92], and highly complex benchmark of CEC 2022 (CEC22) [93] as a single objective-constrained bounded numerical optimization benchmark, each having a specific characteristic. The test functions were shifted and rotated by the values shown in Tables (19-23) in Appendix A to increase the complexity of the problems. The results compared with many optimization algorithms such as MFO FDO, Fox, ANA, PSO, GA, BKA, OOBO [13,16,25,31,35,55,67,94].

Despite the increasing complexity of the tested functions by rotating and shifting, all uni-modal test functions have a single optimum solution, while increasing the dimension will increase the problem difficulty and computation time to reach a globally optimum solution. The test functions F1-F7 are shown in the uni-modal function considered, while the multi-modal function has many local optima, which increases the difficulty of the algorithm to find an optimum solution because of trapping in local optima. The test functions F8-13 multi-modal test functions are considered multi-modal problems for comparison, and the specification of functions, rotation, and shift values of the problems are specified in the table in Appendix Table 20. Composition functions are compounds of many functions with rotation, shifting, and add function bias. These functions are important as case studies because the properties of multi-functions are mixed like real-world problems, and they will show the performance of the algorithm in the exploration and exploitation search capability. The test functions F14-F19 mentioned in Table 21 composite functions had $f_{min}=0$ shown, where $\sigma$ is used to coverage range control of each f(x), and $\lambda$ used for compress and stretch the function, all these are tested with composite functions in the current study. Recently, many competition functions have been provided by high-impact conferences to be used as comparison studies for competing for the performance of optimization algorithms. The 100-Digit functions challenge has 10 hard-solved problems as compound functions from the Society for Industrial and Applied Mathematics (SIAM), the purpose of solving such problem in this paper is to find the optimum solution within a specific time because, in the original paper, there is no time limitation for solve problems [92]. The problems are shown in Table 22 the Hundred-Digit Challenge basics with the range of x values, and dimensions of the test problems.

The CEC22 is also used to show the performance of SHOA. Four groups of test functions included as uni-modal just one function F1, basic multi-modal four functions F2-F5, hybrid multi-modal has only three functions F6-F9, and Composition multi-modal has four F9-F12, they are challenge test functions each having a specific characteristic. Test functions have shift, shuffle, and rotate with a matrix downloaded from a special session of conference link. The change of x values with rotation, shift and shuffling values, for hybrid functions will increase the problem complexity.

Numerical examples with dimensions specified in the tables, each test instance run 30 times. The SHOA runs with a population size set to 15, each nest starts with two solutions as a parent of the nest, parents breeding B eggs, nestling birds feeding by the parent during k generations of algorithm cycles, then best 2 keep for next generation, other birds removed from nest, 500 iterations specified for each turn the SHOA and specified optimization algorithm parameters shown in Table 1. The statical results "Mean "represents the mean value and "Std" is the standard deviation over 30 rounds, the algorithm's extra parameters and specifications are summarized in the Table 1.

The proposed algorithm applies groups and subgroups, the concept designed for multi-modality, but uni-modal test functions are also benchmarked to show the performance of SHOA. The comparison F1, F2, and F7 are median when compared with other algorithms, but in all multi-modal functions have good performance in some show superiority over others. The comparative results mean and standard deviation are shown in Table 2. SHOA outperforms other powerful algorithms in hybrid multi-modal functions.

In Table 3 comparison results for Hundred-Digit challenge problems of SHOA and other algorithms are shown, C01 and C06 are highly complex problems that need more execution time to solve, some algorithms have slow convergence to optimal like ANA, FDO, and PSO takes a lot of time because it has exploited every solution, furthermore PSO fails for C01 and C06 and ANA fails for C06, this proves that ANA and PSO's slow convergence rate during execution. Indeed, results in many test functions like (C01, and C06) show the novel SHOA is more powerful than other algorithms not only at the average value of 30 runs but at other statistical Std values also.

Table 1 Algorithm Specifications

All algorithms run under the same conditions: max iteration = 500
Number of agents in population = 30
Number of rounds = 30
Stop condition = maximum iteration
Extra parameters are specified below:

| | |
|---|---|
| GA | Selection Roulette Wheel<br>Linear combination<br>Crossover arithmetic = 0.8<br>Mutation rate = 0.05 |
| PSO | $C_1, C_2 = (2,2)$<br>Inertia weight = 0.6<br>$r \in [-1,1]$ |
| MFO | Flame size = moth size<br>All dimensions have the same bound<br>a linearly decreases from -1 to -2 |
| FDO | Initial Weight Factor = 0.0<br>r levy flight |
| ANA | $r \in [-1,1]$ |
| SHOA | B = 7<br>k = 50 |

Furthermore, Table 4 shows the comparison result of CEC22 for all problems, SHOA's performance is low compared with other algorithms in Ce01, while in all others it has good performance. Once again, the Wilcoxon rank sum and Fridman test shown in Tables (5-11) demonstrated the statistical performance of SHOA in solving all test problem functions. Two algorithms are compared using the Wilcoxon rank sum test, the Wilcoxon rank sum test a nonparametric statistical test to determine the significant difference between the average of two data samples, is applied. In the Wilcoxon rank sum test, using an index called a p-value of 5%, it is determined whether the superiority of SHOA against any other algorithms is significant from a statistical point of view. While more than one algorithm is used to test for Fridman, any value that is not applicable has a sign (-) in Table (2,3, or 4) Fridman and Wilcoxon tests are not considered for not applicable results.

The Wilcoxon rank sum is reported in Table 5-7. Based on these results, in cases where the p-value is less than 0.05, SHOA has a significant statistical superiority compared to the corresponding algorithm.

In Table 7 the test results show the statistical difference between SHOA with correspondence algorithm specified in, F2, and F7 has p-value > 0.05% while all others are < 0.05, furthers more in Table 6, C4 with ANA, and C7 with FDO have p-value > 0.05, in Table 7 the functions Ce05, and Ce06 with ANA also have p-vales > 0.05. The Fridman statistical test is reported in Tables (8-10). In Table 8 Fridman's results of unimodal functions are presented, MFO takes first place the lowest is the best, and SHOA takes fourth place as it runs median, while MFO is inapplicable to find a solution for F7 under the same conditions which other algorithms found. In Table 9 as all test functions are multi-modal mean ranking of Fridman shows that SHOA takes first place followed by ANA. MFO, FDO, BKA, GA, FOX, OOBO, AND the last PSO.

Hundred-Digit CEC19 test case Fridman results show SHOA taking first ranking place followed by MFO, FDO, ANA, BKA, FOX, OOBO, GA, and PSO. Comparison of Fridman for the last test instances presented in Tables 10 and 11 for both CEC19 and CEC22 respectively, in the overall mean ranking has the lowest value, but in all cases, it performs well even in medium and good performance for all cases.

Furthermore, Figure 7 reports the best solution found during the search space of SHOA with all competitive algorithms selected for Function (F1-F19), the algorithms convergence curve shows smoothly converges to the best solution.

In Figure 8, the 100-digit problems' convergence curve is presented to show the searching space and converge to the optimal solution, some algorithms converge to the optimal solution from the first 25 iterations, while others need more. MFO has fast convergence in C01, C02. SHOA also converges to optimal at the first quarter of iterations. Figure 9 presents convergence curve of CEC22 test instance, some lines appear in some functions meaning multiple algorithms had same value or near value, the line at the top hides others.

The authors of JDE100 [95] run 50 runs for each function with a different initial population but only the best 25 are selected for the final result shown in Table 12, while in SHOA 30 consecutive runs were implemented with different initial populations and all used in the resulted table, and SHOA run on 5e+02 maximum function evaluation, but JDE100 maximum evaluation is 1e+12. In all function comparisons, SHOA performs less mean and std than JDE100. The results are shown in Table 12.

Table 13 shows the performance of SHOA for CEC22 benchmarks for each ( 2, 10, 20 ) dimension, where D = 20 has already been studied but the Table above shows a comparison, there are no application results for F6, F7, F8 in two dimensions because of hybrid functions need more dimensions. The performance of SHOA wouldn't be different with high dimensions when comparing between 10 and 20 dimensions, but overall algorithm performance will be high in small dimensions like D = 2.

Table 14 presents the comparison results of different nesting sizes (4 to 7) and selects samples from the CEC22 instance test cases for D = 10, due to the challenge of achieving an optimal or semi-optimal solution through multiple algorithms at the appropriate time. The results show that with increasing nest size, mean and std with a single model were improved, while multi-modal was complex and difficult to solve; no significant difference was found.

Table 2 Comparison Result of SHOA on 19 Test Instances (Uni , Multi, Composition ) -modals with Algorithms **MFO, FDO, FDO, FOX, ANA, PSO, GA, BKA, OOBO**

| F | SHOA | | MFO | | FDO | | FOX | | ANA | | PSO | | GA | | BKA | | OOBO | |
|---|---|---|---|---|---|---|---|---|---|---|---|---|---|---|---|---|---|---|
| | Mean | STD | Mean | STD | Mean | STD | Mean | STD | Mean | STD | Mean | STD | Mean | STD | Mean | STD | Mean | STD |
| F1 | 1.42E-01 | 7.85E-02 | 3.07E-11 | 3.68E-11 | **1.50E-12** | 7.89E-07 | 5.33E+02 | 5.33E+02 | 1.02E-03 | 2.03E-03 | 7.04E+03 | 1.85E+03 | 3.97E+03 | 1.04E+03 | 5.01E+02 | 1.77E+02 | 5.65E+03 | 1.29E+03 |
| F2 | 4.38E-01 | 5.18E+00 | **1.01E-07** | 1.08E-07 | 1.04E-02 | 1.28E-03 | 9.71E+00 | 1.72E+00 | 3.49E-05 | 2.70E-05 | 5.45E+01 | 4.34E+01 | 1.85E+01 | 3.54E+00 | 4.71E+00 | 1.21E+00 | 1.77E+01 | 2.95E+00 |
| F3 | 1.06E+01 | 9.47E-06 | **4.59E-02** | 6.36E-02 | 4.14E+00 | 1.06E-01 | 1.31E+03 | 4.94E+02 | 1.60E+00 | 1.23E+00 | 1.88E+04 | 1.13E+04 | 6.06E+03 | 1.28E+03 | 8.32E+02 | 2.36E+02 | 4.82E+03 | 8.42E+02 |
| F4 | 8.51E-06 | 9.54E+00 | **0.00E+00** | **0.00E+00** | 9.14E-09 | 7.99E-03 | 9.66E-01 | 1.21E+00 | 1.15E-08 | 6.21E-08 | 1.06E+00 | 1.97E+00 | 2.91E-03 | 1.57E-02 | 8.75E-08 | 2.78E-07 | 7.68E-01 | 6.38E-01 |
| F5 | 2.42E+01 | 1.80E-01 | **6.58E+00** | 1.97E+00 | 6.19E+01 | 4.47E+01 | 2.43E+05 | 1.55E+05 | 7.63E+01 | 2.23E+02 | 4.46E+07 | 0.00E+00 | 1.07E+07 | 4.52E+06 | 6.44E+04 | 4.10E+04 | 4.17E+06 | 3.21E+06 |
| F6 | **3.33E-02** | 2.91E-01 | 4.22E+06 | 7.55E+01 | 4.22E+06 | 4.22E+06 | 4.81E+06 | 9.89E+04 | 4.22E+06 | 3.36E+02 | 5.15E+06 | 1.23E+05 | 5.03E+06 | 7.20E+04 | 4.38E+06 | 4.00E+04 | 4.75E+06 | 8.97E+04 |
| F7 | 7.78E-01 | 2.33E+02 | - | - | 7.77E-01 | 6.22E-01 | 7.22E-01 | 3.18E-01 | 7.47E-01 | **2.77E-01** | 8.44E-01 | 3.66E-01 | - | - | 1.01E+00 | 4.07E-01 | 1.11E+00 | 4.62E-01 |
| F8 | -5.37E+03 | 3.28E+00 | -5.36E+03 | 2.14E+02 | -5.01E+03 | -5.55E+03 | -3.46E+03 | 2.69E+02 | **-2.79E+06** | 2.67E+05 | -2.56E+03 | 2.62E+02 | -2.69E+03 | 5.66E+02 | -4.45E+03 | 3.05E+02 | -3.01E+03 | 2.86E+02 |
| F9 | 1.56E+01 | 1.01E-01 | **8.60E+00** | 5.66E+00 | 2.06E+01 | 1.19E+00 | 3.66E+01 | 4.26E+00 | 2.62E+01 | 3.51E+00 | 1.10E+02 | 1.90E+01 | 2.70E+01 | 2.82E+00 | 3.89E+01 | 6.46E+00 | 7.07E+01 | 9.00E+00 |
| F10 | 3.67E-01 | 7.46E-02 | 2.70E-06 | 1.99E-06 | 7.55E-15 | 7.55E-15 | 2.45E+00 | 8.20E-01 | 5.63E-14 | 2.72E-13 | 1.76E+01 | 1.09E+00 | **4.44E-16** | 0.00E+00 | 7.52E+00 | 9.23E-01 | 1.61E+01 | 1.12E+00 |
| F11 | **2.36E-01** | 5.74E-01 | 5.13E-01 | 6.93E-02 | 5.03E-01 | 5.88E-01 | 5.48E-01 | 1.42E-01 | 4.19E-01 | 6.72E-02 | 8.89E-01 | 8.26E-02 | 4.96E-01 | 8.22E-02 | 4.77E-01 | 9.74E-02 | 4.99E-01 | 8.74E-02 |
| F12 | 8.39E-01 | 1.26E-02 | **9.64E-02** | 4.41E-01 | 5.89E+01 | 6.73E+00 | 3.76E+04 | 5.80E+04 | 3.94E+00 | 4.78E+00 | 3.32E+08 | 2.22E+08 | 3.34E+07 | 1.36E+07 | 2.09E+02 | 1.44E+02 | 1.26E+07 | 1.62E+07 |
| F13 | **4.09E-02** | 2.94E-05 | 4.10E+09 | 7.62E+05 | 4.10E+09 | 4.10E+09 | 1.38E+10 | 4.37E+09 | 4.12E+09 | 8.31E+06 | 5.87E+10 | 1.40E+10 | 4.29E+10 | 7.63E+09 | 7.80E+09 | 9.76E+08 | 2.40E+10 | 6.11E+09 |
| F14 | 8.32E-05 | 5.01E-06 | **4.56E-16** | 6.69E-16 | 2.36E-15 | 1.19E-14 | 2.64E-02 | 5.56E-03 | 2.73E-09 | 1.46E-08 | 4.33E-01 | 3.90E-01 | 9.73E-03 | 1.93E-03 | 3.04E-03 | 1.26E-03 | 2.80E-02 | 6.10E-03 |
| F15 | 1.54E-05 | 1.69E-03 | 7.47E-10 | 5.59E-10 | **8.88E-16** | 7.77E-16 | 2.11E-01 | 2.27E-02 | 1.72E-08 | 1.49E-08 | 8.20E-02 | 2.10E-02 | 1.20E-01 | 1.70E-02 | 5.41E-02 | 1.18E-02 | 1.94E-01 | 1.55E-02 |
| F16 | 5.35E-03 | 3.36E-01 | 2.35E-08 | 1.58E-08 | **4.33E-15** | 1.67E-15 | 1.02E+00 | 8.63E-03 | 4.20E-06 | 2.85E-06 | 1.06E+00 | 1.66E-02 | 9.49E-01 | 3.40E-02 | 7.62E-01 | 8.73E-02 | 1.03E+00 | 8.46E-03 |
| F17 | **4.53E+00** | 8.94E-02 | 2.38E+01 | 7.94E-02 | 2.40E+01 | 2.37E+01 | 2.40E+01 | 1.88E-01 | 2.38E+01 | 5.36E-01 | 3.23E+01 | 6.90E+00 | 2.44E+01 | 4.90E-01 | 2.37E+01 | 6.19E-02 | 2.39E+01 | 1.38E-01 |
| F18 | **3.21E+00** | 6.30E-02 | 2.24E+02 | 3.93E-03 | 2.24E+02 | 2.24E+02 | 2.24E+02 | 1.66E-02 | 2.24E+02 | 6.62E-03 | 2.24E+02 | 6.03E-02 | 2.24E+02 | 2.14E-02 | 2.24E+02 | 4.69E-03 | 2.24E+02 | 9.34E-03 |
| F19 | **3.80E+00** | 0.00E+00 | 3.15E+01 | 6.51E-03 | 3.15E+01 | 3.15E+01 | 3.20E+01 | 2.04E-01 | 3.15E+01 | 2.19E-02 | 4.72E+01 | 1.48E+01 | 3.35E+01 | 8.66E-01 | 3.18E+01 | 9.61E-02 | 3.19E+01 | 1.58E-01 |

Table 3 Comparison Result of SHOA on Hundred- Digit CEC 2019 Test Instances with Algorithms MFO, FDO, FDO, FOX, ANA, PSO, GA, BKA, OOBO

| F | SHOA | | MFO | | FDO | | FOX | | ANA | | PSO | | GA | | BKA | | OOBO | |
|---|---|---|---|---|---|---|---|---|---|---|---|---|---|---|---|---|---|---|
| | Mean | STD | Mean | STD | Mean | STD | Mean | STD | Mean | STD | Mean | STD | Mean | STD | Mean | STD | Mean | STD |
| C1 | **2.72E-01** | 5.29E-06 | 2.04E+05 | 3.02E+05 | 4.59E+03 | 2.07E+04 | 2.71E+05 | 3.57E+05 | 2.15E+06 | 2.08E+06 | - | - | 8.05E+03 | 3.97E+03 | 2.85E+06 | 2.38E+06 | 1.11E+07 | 6.99E+06 |
| C2 | **3.00E+00** | 2.41E-01 | 4.00E+00 | 0.00E+00 | 4.00E+00 | 2.58E-05 | 4.78E+00 | 1.16E+00 | 4.00E+00 | 3.98E-12 | 3.89E+02 | 2.98E+02 | 4.68E+00 | 3.70E-01 | 5.35E+00 | 1.33E+00 | 4.00E+00 | 3.74E-04 |
| C3 | **3.13E+00** | 6.86E+00 | 1.37E+01 | 4.88E-13 | 1.37E+01 | 9.89E-08 | 1.37E+01 | 2.45E-05 | 1.37E+01 | 1.76E-10 | 1.37E+01 | 1.29E-03 | 1.37E+01 | 4.52E-04 | 1.37E+01 | 4.67E-06 | 1.37E+01 | 1.66E-04 |
| C4 | 4.36E+01 | 3.79E-02 | 2.94E+01 | 1.11E+01 | **1.17E+00** | 8.47E-02 | 8.01E+03 | 2.71E+03 | 4.25E+01 | 9.90E+00 | 2.47E+04 | 8.39E+03 | 1.29E+04 | 3.38E+03 | 6.56E+02 | 2.30E+02 | 6.33E+03 | 1.93E+03 |
| C5 | **1.33E-01** | 7.06E-01 | 1.12E+00 | 1.31E-01 | 2.14E+00 | 8.72E-02 | 3.62E+00 | 4.29E-01 | 1.20E+00 | 8.45E-02 | 6.68E+00 | 0.00E+00 | 4.30E+00 | 4.98E-01 | 2.16E+00 | 6.94E-02 | 3.49E+00 | 5.10E-01 |
| C6 | **7.97E+00** | 4.15E+00 | 1.21E+01 | 8.36E-01 | 1.21E+01 | 6.10E-01 | 1.21E+01 | 6.12E-01 | - | - | - | - | 1.24E+01 | 7.97E-01 | 1.18E+01 | 8.18E-01 | 1.22E+01 | 5.68E-01 |
| C7 | 1.19E+02 | 3.73E-01 | **1.06E+02** | 3.48E+00 | 1.22E+02 | 1.43E+01 | 4.52E+02 | 7.22E+01 | 1.16E+02 | 8.17E+00 | 6.01E+02 | 6.04E+01 | 4.75E+02 | 3.52E+01 | 2.36E+02 | 2.70E+01 | 3.72E+02 | 3.68E+01 |
| C8 | **2.52E+00** | 6.71E-03 | 4.96E+00 | 5.77E-01 | 5.14E+00 | 9.00E-01 | 5.85E+00 | 4.76E-01 | 5.53E+00 | 5.12E-01 | 8.25E+00 | 5.59E-01 | 6.33E+00 | 4.19E-01 | 6.07E+00 | 3.98E-01 | 6.28E+00 | 2.74E-01 |
| C9 | **1.02E+00** | 8.88E-16 | 2.00E+00 | 9.11E-12 | 2.00E+00 | 1.75E-06 | 2.95E+02 | 1.43E+02 | 2.00E+00 | 5.52E-04 | 8.00E+02 | 2.08E+02 | 2.52E+02 | 5.89E+01 | 2.74E+01 | 1.04E+01 | 2.41E+02 | 6.19E+01 |
| C10 | **1.72E+00** | 0.00E+00 | 2.72E+00 | 4.44E-16 | 2.72E+00 | 4.44E-16 | 2.72E+00 | 4.44E-16 | 2.72E+00 | 4.44E-16 | 2.72E+00 | 4.44E-16 | 2.72E+00 | 4.44E-16 | 2.72E+00 | 4.44E-16 | 2.72E+00 | 4.44E-16 |

Table 4 Comparison Result of SHOA on CEC 2022 Test Instances with Algorithms MFO, FDO, FDO, FOX, ANA, PSO, GA, BKA, OOBO

| F | SHOA | | MFO | | FDO | | FOX | | ANA | | PSO | | GA | | BKA | | OOBO | |
|---|---|---|---|---|---|---|---|---|---|---|---|---|---|---|---|---|---|---|
| | Mean | STD | Mean | STD | Mean | STD | Mean | STD | Mean | STD | Mean | STD | Mean | STD | Mean | STD | Mean | STD |
| Ce1 | 1.84E+03 | 8.95E+01 | 3.44E+04 | 6.54E+03 | 1.83E+03 | 2.44E+03 | 2.45E+04 | 3.34E+03 | **3.78E+01** | 2.50E+01 | 2.01E+06 | 7.61E+06 | 5.57E+04 | 1.18E+04 | 1.40E+04 | 2.54E+03 | 3.62E+04 | 5.14E+03 |
| Ce2 | **4.51E+02** | 1.42E-02 | 4.07E+03 | 6.86E+02 | 5.67E+00 | 3.75E+00 | 2.61E+03 | 3.97E+02 | 3.25E+01 | 1.86E+01 | 6.35E+03 | 1.02E+03 | 4.32E+03 | 4.20E+02 | 1.27E+03 | 2.10E+02 | 4.38E+03 | 9.01E+02 |
| Ce3 | 5.39E-02 | 1.21E+01 | 1.59E+00 | 1.70E-01 | 1.99E-05 | 6.34E-05 | 1.02E+00 | 1.30E-01 | **4.33E-06** | 4.99E-06 | 2.14E+00 | 3.09E-01 | 1.80E+00 | 1.53E-01 | 3.54E-01 | 7.92E-02 | 1.74E+00 | 2.85E-01 |
| Ce4 | 9.97E+01 | 4.27E-01 | 2.11E+02 | 1.41E+01 | **6.71E+01** | 2.40E+01 | 1.73E+02 | 1.82E+01 | 1.32E+02 | 1.16E+01 | 2.31E+02 | 1.86E+01 | 1.91E+02 | 1.29E+01 | 1.62E+02 | 8.32E+00 | 2.26E+02 | 1.63E+01 |
| Ce5 | 1.87E+00 | 1.11E+01 | 1.35E+01 | 2.21E+00 | 1.45E+00 | 9.24E-01 | 8.07E+00 | 1.85E+00 | 1.88E+00 | 7.55E-01 | 1.78E+01 | 0.00E+00 | 8.53E+00 | 1.26E+00 | 4.98E+00 | 1.16E+00 | 1.56E+00 | 2.16E+00 |
| Ce6 | **7.13E+01** | 1.52E+00 | 9.38E+08 | 3.74E+08 | 1.48E+04 | 5.08E+04 | 1.06E+09 | 1.25E+08 | 8.12E+02 | 2.90E+03 | 2.27E+03 | 8.67E+02 | 1.09E+03 | 3.04E+02 | 2.21E+02 | 5.19E+01 | 1.02E+03 | 3.22E+02 |
| Ce7 | 2.74E+01 | 3.18E+00 | 5.02E+02 | 1.72E+02 | 4.06E+01 | 1.32E+01 | 1.47E+03 | 6.82E+02 | 3.43E+01 | 3.52E+00 | 7.19E+01 | 1.43E+01 | 4.77E+01 | 4.69E+00 | 3.06E+01 | 2.77E+00 | 4.58E+01 | 4.72E+00 |
| Ce8 | **2.39E+01** | 3.72E+00 | 3.36E+02 | 1.11E+02 | 2.81E+01 | 5.04E+00 | 4.04E+02 | 1.71E+02 | 2.94E+01 | 1.39E+00 | 6.38E+01 | 1.32E+01 | 4.32E+01 | 3.41E+00 | 2.63E+01 | 1.70E+00 | 3.62E+01 | 2.01E+00 |
| Ce9 | **1.91E+02** | 9.48E-02 | 3.70E+02 | 3.97E+01 | **1.77E+02** | 1.54E+00 | 7.67E+02 | 3.16E+02 | 1.78E+02 | 1.79E+00 | 1.31E+03 | 4.94E+02 | 7.81E+02 | 1.45E+02 | 2.47E+02 | 1.68E+01 | 4.10E+02 | 5.11E+01 |
| Ce10 | 1.01E+02 | 3.03E+01 | 7.41E+02 | 6.39E+02 | 2.06E+01 | 4.37E+01 | 2.00E+03 | 8.83E+02 | **4.19E+01** | 5.07E+01 | 5.48E+03 | 3.03E+02 | 1.59E+03 | 8.85E+02 | 5.14E+02 | 2.68E+02 | 3.44E+02 | 3.26E+02 |
| Ce11 | **1.33E+02** | 2.57E+00 | 2.86E+03 | 5.99E+02 | 5.48E+02 | 5.61E+02 | 5.86E+03 | 9.92E+02 | 4.60E+02 | 4.23E+02 | 7.85E+03 | 7.27E+02 | 4.28E+03 | 6.19E+02 | 1.54E+03 | 2.88E+02 | 3.04E+03 | 5.92E+02 |
| Ce12 | **2.26E+02** | 0.00E+00 | 4.15E+02 | 2.37E+01 | 2.79E+02 | 5.04E+01 | 5.72E+02 | 6.44E+01 | 2.87E+02 | 5.65E+00 | 7.08E+02 | 7.19E+01 | 3.43E+02 | 1.46E+02 | 3.52E+02 | 1.11E+01 | 4.38E+02 | 2.70E+01 |

Table 5 Wilcoxon Ranking Sum Test SHOA on 19 Different Modal test instances against Algorithms
SHOA vs (MFO, FDO, FDO, FOX, ANA, PSO, GA, BKA, OOBO)

| F | MFO | FDO | FOX | ANA | PSO | GA | BKA | OOBO |
|---|---|---|---|---|---|---|---|---|
| F1 | ~ 0.01 | ~ 0.01 | ~ 0.01 | ~ 0.01 | ~ 0.01 | ~ 0.01 | ~ 0.01 | ~ 0.01 |
| F2 | ~ 0.01 | **0.673** | ~ 0.01 | ~ 0.01 | ~ 0.01 | ~ 0.01 | ~ 0.01 | ~ 0.01 |
| F3 | ~ 0.01 | ~ 0.01 | ~ 0.01 | ~ 0.01 | ~ 0.01 | ~ 0.01 | ~ 0.01 | ~ 0.01 |
| F4 | ~ 0.01 | ~ 0.01 | ~ 0.01 | ~ 0.01 | ~ 0.01 | ~ 0.01 | ~ 0.01 | ~ 0.01 |
| F5 | ~ 0.01 | ~ 0.01 | ~ 0.01 | 0.013 | ~ 0.01 | ~ 0.01 | ~ 0.01 | ~ 0.01 |
| F6 | ~ 0.01 | ~ 0.01 | ~ 0.01 | ~ 0.01 | ~ 0.01 | ~ 0.01 | ~ 0.01 | ~ 0.01 |
| F7 | - | **0.797** | **0.861** | 0.491 | 0.01 | - | **0.861** | 0.003 |
| F8 | ~ 0.01 | ~ 0.01 | ~ 0.01 | ~ 0.01 | ~ 0.01 | ~ 0.01 | ~ 0.01 | ~ 0.01 |
| F9 | ~ 0.01 | ~ 0.01 | ~ 0.01 | ~ 0.01 | ~ 0.01 | ~ 0.01 | ~ 0.01 | ~ 0.01 |
| F10 | ~ 0.01 | ~ 0.01 | ~ 0.01 | ~ 0.01 | ~ 0.01 | ~ 0.01 | ~ 0.01 | ~ 0.01 |
| F11 | ~ 0.01 | ~ 0.01 | ~ 0.01 | ~ 0.01 | ~ 0.01 | ~ 0.01 | ~ 0.01 | ~ 0.01 |
| F12 | ~ 0.01 | ~ 0.01 | ~ 0.01 | ~ 0.01 | ~ 0.01 | ~ 0.01 | ~ 0.01 | ~ 0.01 |
| F13 | ~ 0.01 | ~ 0.01 | ~ 0.01 | ~ 0.01 | ~ 0.01 | ~ 0.01 | ~ 0.01 | ~ 0.01 |
| F14 | ~ 0.01 | ~ 0.01 | ~ 0.01 | ~ 0.01 | ~ 0.01 | ~ 0.01 | ~ 0.01 | ~ 0.01 |
| F15 | ~ 0.01 | ~ 0.01 | ~ 0.01 | ~ 0.01 | ~ 0.01 | ~ 0.01 | ~ 0.01 | ~ 0.01 |
| F16 | ~ 0.01 | ~ 0.01 | ~ 0.01 | ~ 0.01 | ~ 0.01 | ~ 0.01 | ~ 0.01 | ~ 0.01 |
| F17 | ~ 0.01 | ~ 0.01 | ~ 0.01 | ~ 0.01 | ~ 0.01 | ~ 0.01 | ~ 0.01 | ~ 0.01 |
| F18 | ~ 0.01 | ~ 0.01 | ~ 0.01 | ~ 0.01 | ~ 0.01 | ~ 0.01 | ~ 0.01 | ~ 0.01 |
| F19 | ~ 0.01 | ~ 0.01 | ~ 0.01 | ~ 0.01 | ~ 0.01 | ~ 0.01 | ~ 0.01 | ~ 0.01 |

Table 6 Wilcoxon Ranking Sum Test SHOA on Hundred Digit CEC 2019 test instances against Algorithms
SHOA vs (MFO, FDO, FDO, FOX, ANA, PSO, GA, BKA, OOBO)

| F | MFO | FDO | FOX | ANA | PSO | GA | BKA | OOBO |
|---|---|---|---|---|---|---|---|---|
| C1 | ~ 0.01 | ~ 0.01 | ~ 0.01 | ~ 0.01 | - | ~ 0.01 | ~ 0.01 | ~ 0.01 |
| C2 | ~ 0.01 | ~ 0.01 | ~ 0.01 | ~ 0.01 | ~ 0.01 | ~ 0.01 | ~ 0.01 | ~ 0.01 |
| C3 | ~ 0.01 | ~ 0.01 | ~ 0.01 | ~ 0.01 | ~ 0.01 | ~ 0.01 | ~ 0.01 | ~ 0.01 |
| C4 | ~ 0.01 | ~ 0.01 | ~ 0.01 | **0.629** | ~ 0.01 | ~ 0.01 | ~ 0.01 | ~ 0.01 |
| C5 | ~ 0.01 | ~ 0.01 | ~ 0.01 | ~ 0.01 | ~ 0.01 | ~ 0.01 | ~ 0.01 | ~ 0.01 |
| C6 | ~ 0.01 | ~ 0.01 | ~ 0.01 | ~ 0.01 | - | ~ 0.01 | ~ 0.01 | ~ 0.01 |
| C7 | ~ 0.01 | 0.797 | ~ 0.01 | 0.014 | ~ 0.01 | ~ 0.01 | ~ 0.01 | ~ 0.01 |
| C8 | ~ 0.01 | ~ 0.01 | ~ 0.01 | ~ 0.01 | ~ 0.01 | ~ 0.01 | ~ 0.01 | ~ 0.01 |
| C9 | ~ 0.01 | ~ 0.01 | ~ 0.01 | ~ 0.01 | ~ 0.01 | ~ 0.01 | ~ 0.01 | ~ 0.01 |
| C10 | ~ 0.01 | ~ 0.01 | ~ 0.01 | ~ 0.01 | ~ 0.01 | ~ 0.01 | ~ 0.01 | ~ 0.01 |

Table 7 Wilcoxon Ranking Sum Test SHOA on CEC 2022 test instances against Algorithms
SHOA vs (MFO, FDO, FDO, FOX, ANA, PSO, GA, BKA, OOBO)

| F | MFO | FDO | FOX | ANA | PSO | GA | BKA | OOBO |
|---|---|---|---|---|---|---|---|---|
| Ce1 | ~ 0.01 | **0.094** | ~ 0.01 | ~ 0.01 | ~ 0.01 | ~ 0.01 | ~ 0.01 | ~ 0.01 |
| Ce2 | ~ 0.01 | ~ 0.01 | ~ 0.01 | ~ 0.01 | ~ 0.01 | ~ 0.01 | ~ 0.01 | ~ 0.01 |
| Ce3 | ~ 0.01 | ~ 0.01 | ~ 0.01 | ~ 0.01 | ~ 0.01 | ~ 0.01 | ~ 0.01 | ~ 0.01 |
| Ce4 | ~ 0.01 | ~ 0.01 | ~ 0.01 | ~ 0.01 | ~ 0.01 | ~ 0.01 | ~ 0.01 | ~ 0.01 |
| Ce5 | ~ 0.01 | 0.011 | ~ 0.01 | **0.975** | ~ 0.01 | ~ 0.01 | ~ 0.01 | ~ 0.01 |
| Ce6 | ~ 0.01 | ~ 0.01 | ~ 0.01 | **0.558** | ~ 0.01 | ~ 0.01 | ~ 0.01 | ~ 0.01 |
| Ce7 | ~ 0.01 | ~ 0.01 | ~ 0.01 | ~ 0.01 | ~ 0.01 | ~ 0.01 | ~ 0.01 | ~ 0.01 |
| Ce8 | ~ 0.01 | ~ 0.01 | ~ 0.01 | ~ 0.01 | ~ 0.01 | ~ 0.01 | ~ 0.01 | ~ 0.01 |
| Ce9 | ~ 0.01 | ~ 0.01 | ~ 0.01 | ~ 0.01 | ~ 0.01 | ~ 0.01 | ~ 0.01 | ~ 0.01 |
| Ce10 | ~ 0.01 | ~ 0.01 | ~ 0.01 | 0.01 | ~ 0.01 | ~ 0.01 | ~ 0.01 | ~ 0.01 |
| Ce11 | ~ 0.01 | ~ 0.01 | ~ 0.01 | ~ 0.01 | ~ 0.01 | ~ 0.01 | ~ 0.01 | ~ 0.01 |
| Ce12 | ~ 0.01 | ~ 0.01 | ~ 0.01 | ~ 0.01 | ~ 0.01 | 0.021 | ~ 0.01 | ~ 0.01 |

Table 8 Fridman Mean Ranking SHOA on 7 single modal test instances against Algorithms MFO, FDO, FDO, FOX, ANA, PSO, GA, BKA, OOBO

| F | SHOA | MFO | FDO | FOX | ANA | PSO | GA | BKA | OOBO |
|---|------|-----|-----|-----|-----|-----|-----|-----|------|
| F1 | 4.00 | 1.17 | 2.00 | 5.60 | 2.83 | 8.67 | 7.27 | 5.40 | 8.07 |
| F2 | 3.73 | 1.00 | 3.30 | 5.97 | 2.13 | 9.00 | 7.57 | 4.87 | 7.43 |
| F3 | 3.93 | 1.23 | 2.00 | 5.83 | 2.83 | 8.93 | 7.90 | 5.17 | 7.17 |
| F4 | 5.23 | 1.87 | 5.73 | 8.07 | 2.25 | 7.70 | 2.05 | 4.03 | 8.07 |
| F5 | 3.03 | 1.20 | 3.57 | 5.93 | 2.20 | 9.00 | 7.80 | 5.07 | 7.20 |
| F6 | 1.00 | 3.00 | 3.00 | 6.70 | 3.00 | 8.83 | 8.10 | 5.00 | 6.37 |
| F7 | 3.67 | - | 3.70 | 3.53 | 3.80 | 4.80 | - | 3.57 | 4.93 |
| Mean Rank | 24.6 | 9.47 | 23.3 | 41.63 | 19.05 | 56.93 | 40.68 | 33.10 | 49.23 |
|  | 4 | 1 | 3 | 7 | 2 | 9 | 6 | 5 | 8 |

Table 9 Fridman Mean Ranking SHOA on 14 Different multi-modal test instances against Algorithms MFO, FDO, FDO, FOX, ANA, PSO, GA, BKA, OOBO

| F | SHOA | MFO | FDO | FOX | ANA | PSO | GA | BKA | OOBO |
|---|------|-----|-----|-----|-----|-----|-----|-----|------|
| F8 | 2.62 | 2.73 | 4.02 | 6.13 | 1.00 | 8.47 | 8.03 | 4.63 | 7.37 |
| F9 | 2.23 | 1.30 | 2.80 | 6.37 | 4.43 | 8.93 | 4.50 | 6.37 | 8.07 |
| F10 | 5.00 | 4.00 | 2.57 | 6.00 | 2.43 | 8.90 | 1.00 | 7.00 | 8.10 |
| F11 | 1.03 | 5.20 | 6.73 | 5.73 | 3.03 | 8.90 | 4.90 | 4.57 | 4.90 |
| F12 | 2.10 | 1.03 | 4.00 | 5.90 | 2.90 | 9.00 | 7.77 | 5.10 | 7.20 |
| F13 | 1.00 | 2.50 | 2.50 | 6.03 | 4.00 | 8.77 | 8.13 | 5.10 | 6.97 |
| F14 | 4.00 | 1.13 | 2.10 | 7.50 | 2.77 | 8.90 | 6.00 | 5.00 | 7.60 |
| F15 | 4.00 | 2.03 | 1.03 | 8.70 | 2.93 | 5.83 | 6.97 | 5.20 | 8.30 |
| F16 | 4.00 | 2.00 | 1.00 | 7.23 | 3.00 | 8.97 | 5.97 | 5.03 | 7.80 |
| F17 | 1.00 | 3.70 | 4.50 | 6.67 | 3.55 | 8.97 | 7.57 | 3.32 | 5.73 |
| F18 | 1.00 | 7.00 | 7.00 | 4.00 | 2.00 | 7.00 | 7.00 | 7.00 | 3.00 |
| F19 | 1.00 | 2.95 | 2.95 | 6.50 | 3.10 | 9.00 | 8.00 | 5.57 | 5.93 |
| Mean Rank | 28.98 | 35.58 | 41.20 | 76.77 | 35.15 | 101.63 | 75.83 | 63.88 | 80.97 |
| Rank | 1 | 3 | 4 | 7 | 2 | 9 | 6 | 5 | 8 |

Table 10 Fridman Mean Ranking SHOA on Hundred CEC 2019 test instances against Algorithms MFO, FDO, FDO, FOX, ANA, PSO, GA, BKA, OOBO

| F | SHOA | MFO | FDO | FOX | ANA | PSO | GA | BKA | OOBO |
|---|------|-----|-----|-----|-----|-----|-----|-----|------|
| C01 | 1.03 | 4.60 | 2.10 | 4.23 | 6.53 | # | 3.20 | 6.47 | 7.83 |
| C02 | 1.00 | 3.00 | 3.00 | 6.53 | 3.00 | 9.00 | 7.07 | 7.40 | 5.00 |
| C03 | 1.00 | 4.00 | 4.00 | 7.00 | 4.00 | 9.00 | 4.00 | 4.00 | 8.00 |
| C04 | 3.45 | 2.37 | 1.00 | 6.77 | 3.18 | 8.90 | 7.90 | 5.00 | 6.43 |
| C05 | 1.00 | 2.25 | 4.37 | 6.77 | 2.75 | 8.90 | 7.73 | 4.63 | 6.60 |
| C06 | 2.00 | 5.62 | 5.58 | 5.37 | 1.00 | 8.93 | 6.32 | 4.52 | 5.67 |
| C07 | 3.17 | 1.22 | 2.95 | 7.30 | 2.67 | 8.87 | 7.63 | 5.00 | 6.20 |
| C08 | 1.00 | 2.95 | 3.57 | 5.00 | 4.02 | 9.00 | 6.90 | 5.97 | 6.60 |
| C09 | 1.00 | 3.00 | 3.00 | 7.47 | 3.00 | 8.93 | 6.80 | 5.03 | 6.77 |
| C10 | 1.00 | 7.00 | 7.00 | 3.00 | 7.00 | 3.00 | 7.00 | 7.00 | 3.00 |
| Mean Rank | 15.65 | 36.00 | 36.57 | 59.43 | 37.15 | 74.53 | 64.55 | 55.02 | 62.10 |
| Rank | 1 | 2 | 3 | 6 | 4 | 9 | 8 | 5 | 7 |

Table 11 Fridman Mean Ranking SHOA on CEC22 test instances against Algorithms MFO, FDO, FDO, FOX, ANA, PSO, GA, BKA, OOBO

| F | SHOA | MFO | FDO | FOX | ANA | PSO | GA | BKA | OOBO |
|---|------|-----|-----|-----|-----|-----|-----|-----|------|
| Ce01 | 2.73 | 6.40 | 2.27 | 5.13 | 1.00 | 9.00 | 7.87 | 4.00 | 6.60 |
| Ce02 | 3.00 | 6.77 | 1.00 | 5.07 | 2.00 | 8.97 | 7.07 | 4.00 | 7.13 |
| Ce03 | 3.00 | 6.57 | 1.13 | 5.03 | 1.87 | 8.60 | 7.47 | 4.00 | 7.33 |
| Ce04 | 1.90 | 7.30 | 1.17 | 4.90 | 3.03 | 8.40 | 5.95 | 4.32 | 8.03 |
| Ce05 | 2.23 | 7.33 | 1.73 | 5.33 | 2.07 | 8.67 | 5.65 | 4.02 | 7.97 |
| Ce06 | 1.87 | 8.37 | 4.57 | 8.63 | 2.00 | 6.40 | 5.13 | 3.20 | 4.83 |
| Ce07 | 1.23 | 8.13 | 3.98 | 8.87 | 3.05 | 6.93 | 5.37 | 2.30 | 5.13 |
| Ce08 | 1.40 | 8.43 | 2.87 | 8.57 | 3.53 | 7.00 | 5.93 | 2.30 | 4.97 |
| Ce09 | 3.00 | 5.30 | 1.37 | 7.37 | 1.63 | 8.70 | 7.77 | 4.00 | 5.87 |
| Ce10 | 2.63 | 5.67 | 1.42 | 7.13 | 1.95 | 9.00 | 6.97 | 5.37 | 4.87 |
| Ce11 | 1.47 | 5.37 | 2.40 | 7.87 | 2.30 | 8.97 | 7.00 | 3.90 | 5.73 |
| Ce12 | 1.50 | 5.77 | 2.62 | 8.02 | 3.38 | 8.92 | 4.00 | 4.50 | 6.30 |
| Mean Rank | 25.97 | 81.40 | 26.52 | 81.92 | 27.82 | 99.55 | 76.17 | 45.90 | 74.77 |
| Rank | 1 | 7 | 2 | 8 | 3 | 9 | 6 | 4 | 5 |

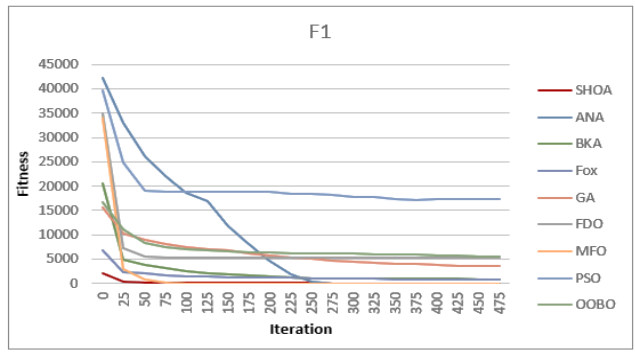
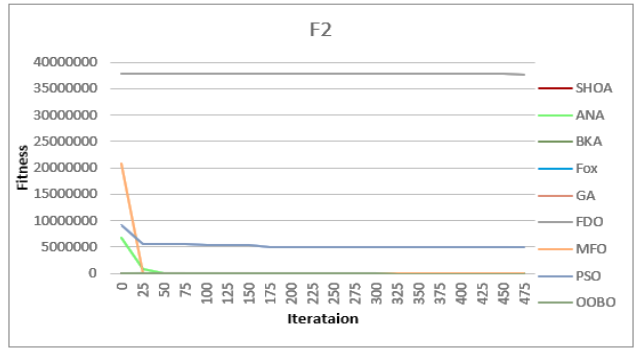
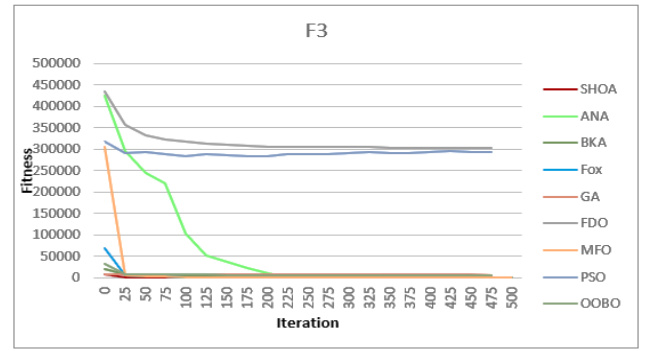
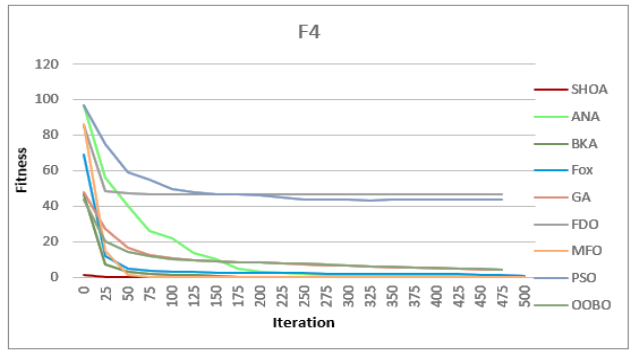
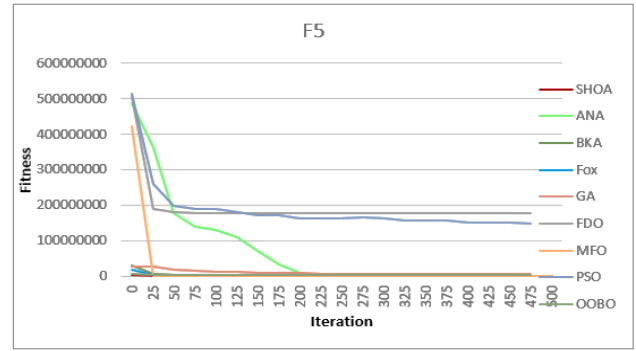
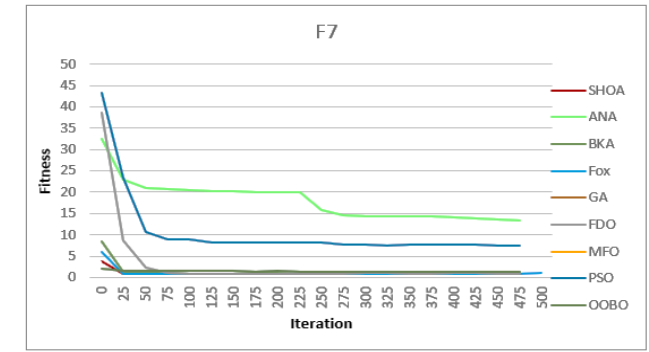
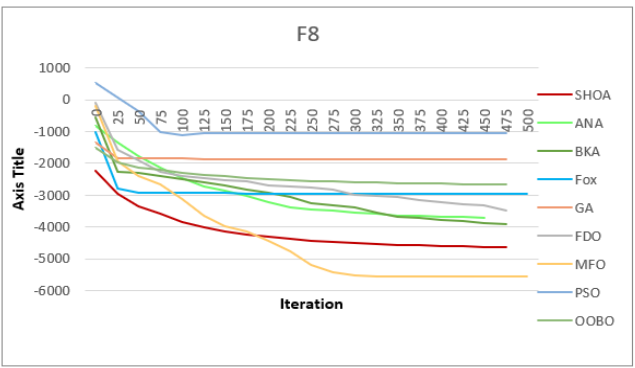
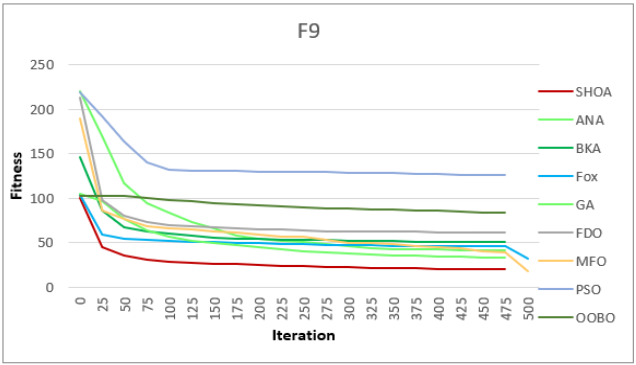
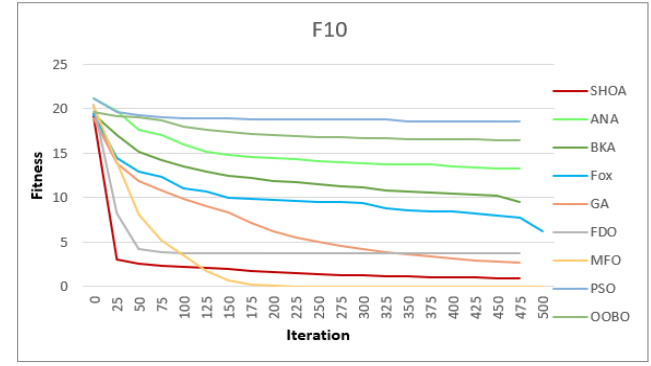

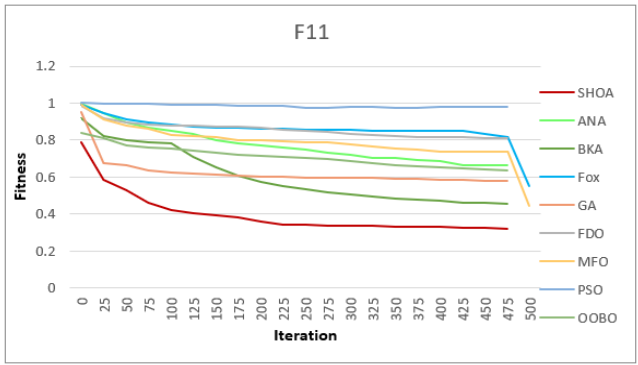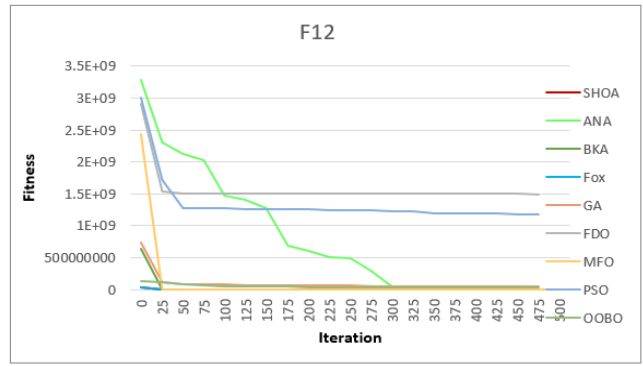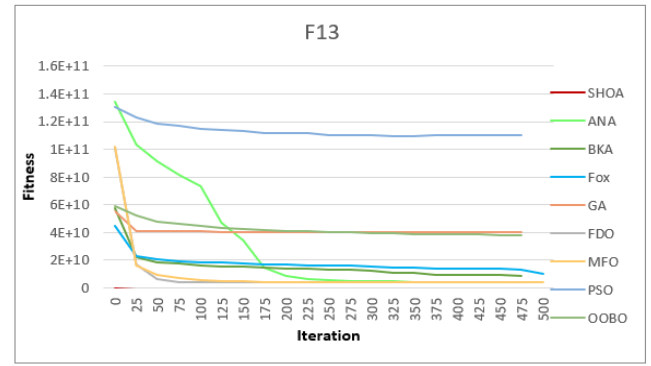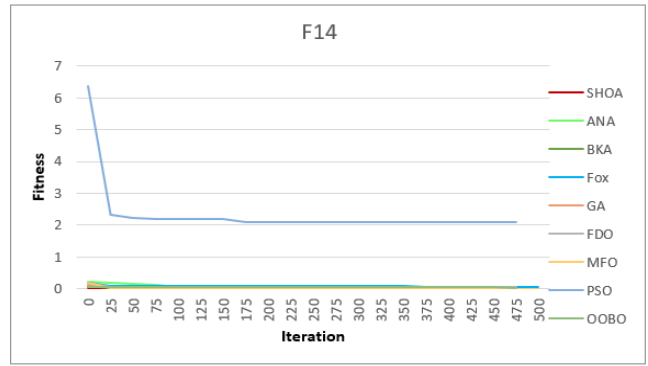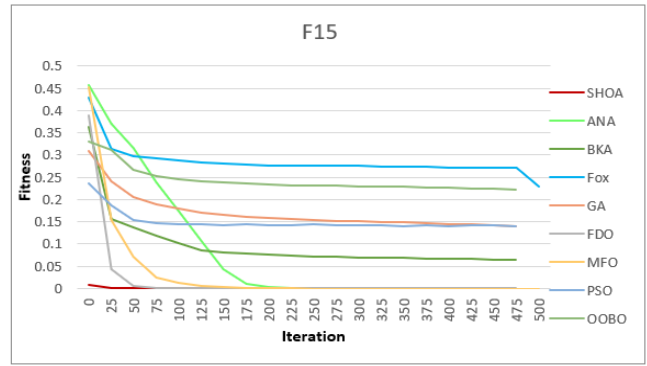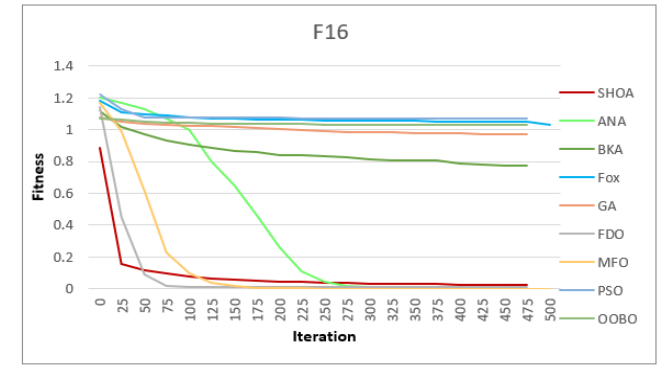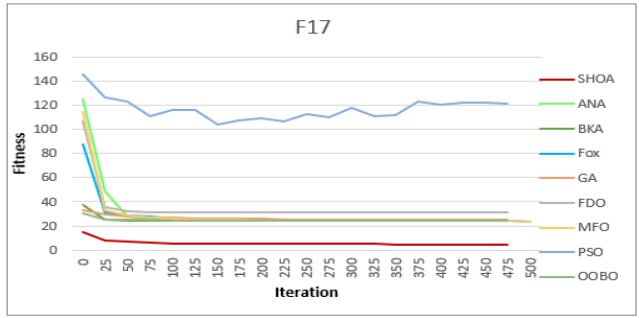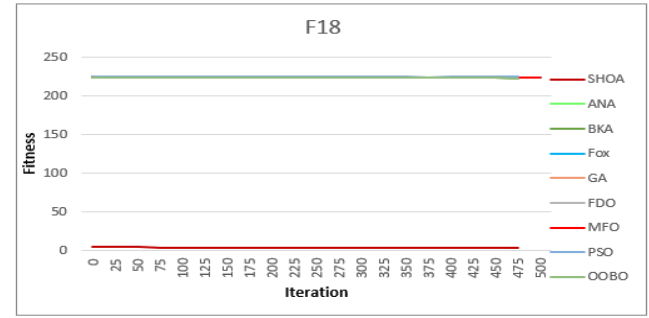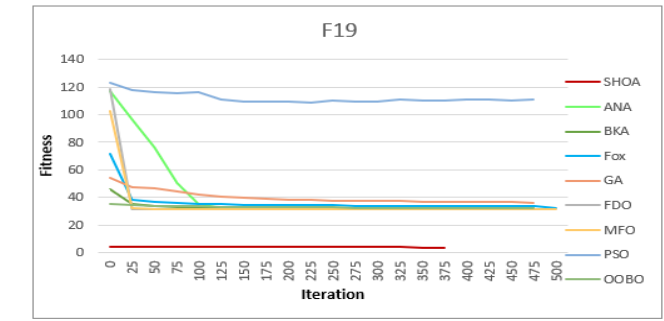

Figure 7 Convergence Curve F1-F19

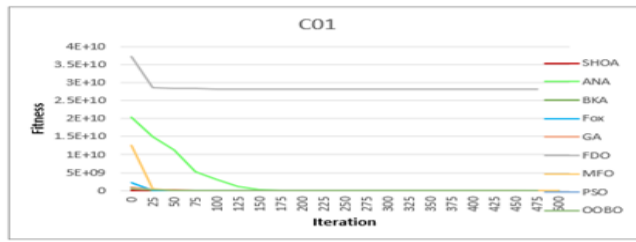 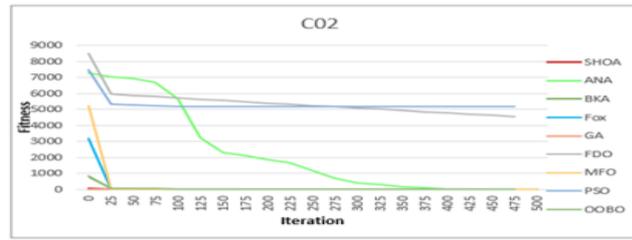 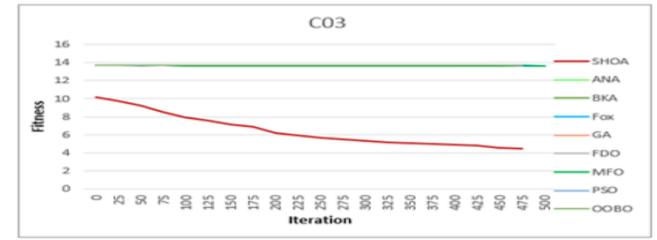
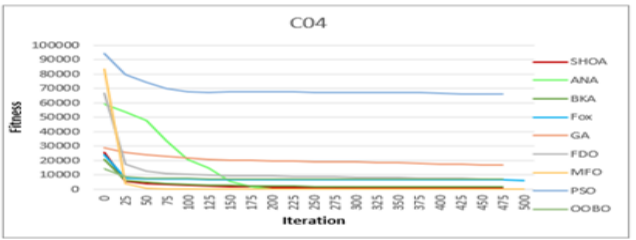 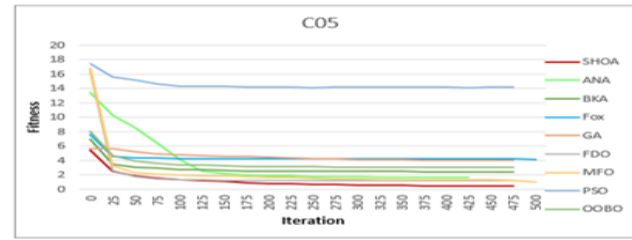 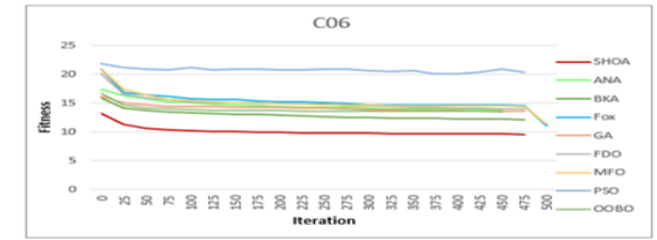
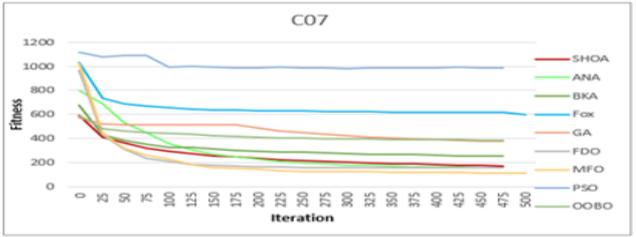 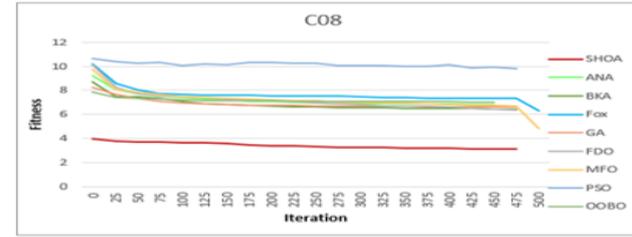 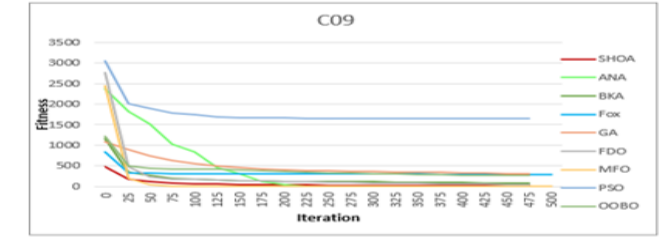

*Figure 8 Convergence Curve CEC 2019*

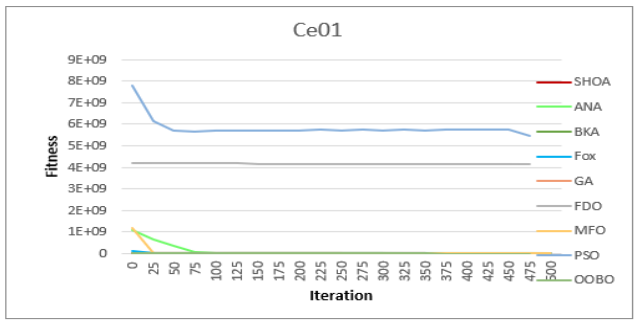 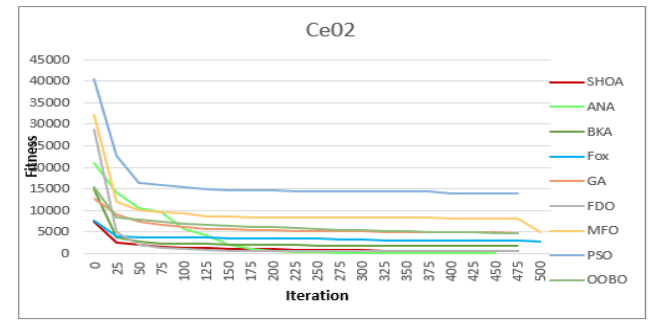 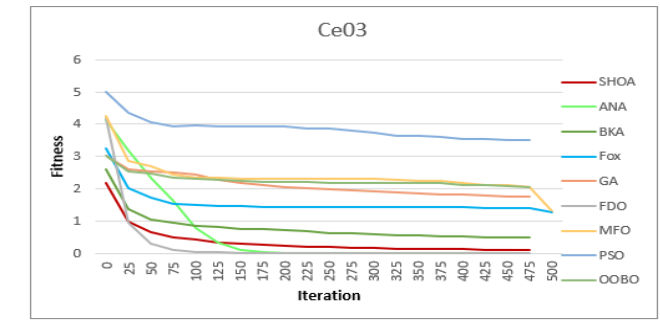

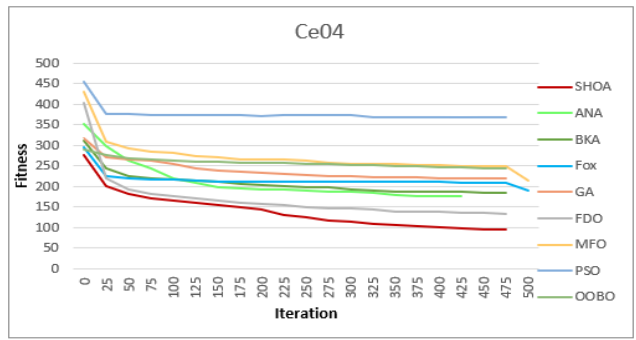
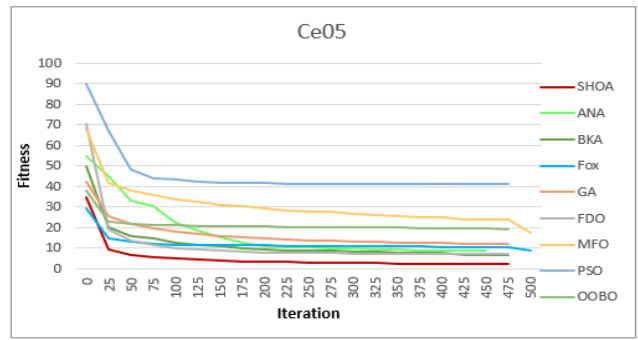
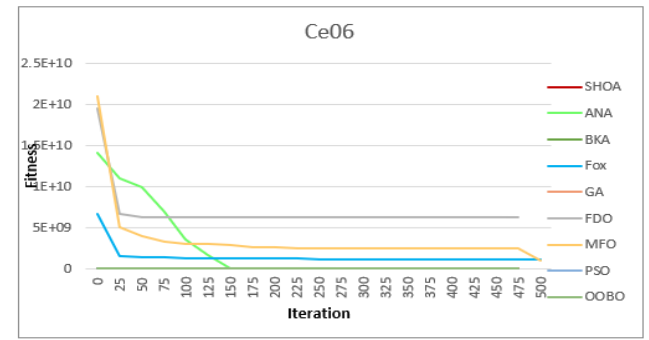
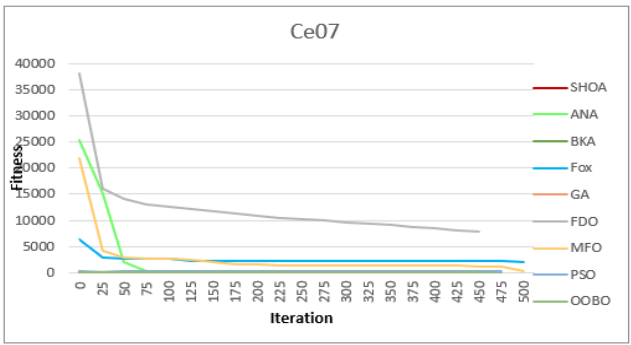
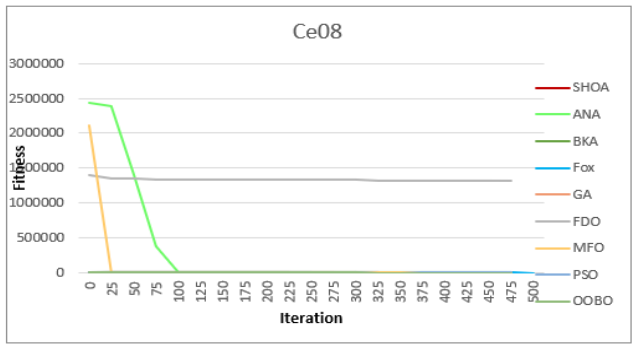
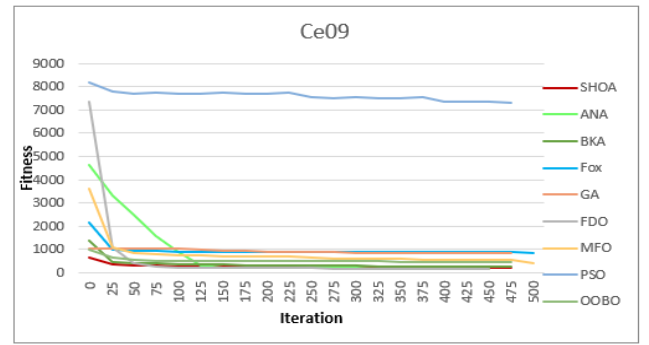
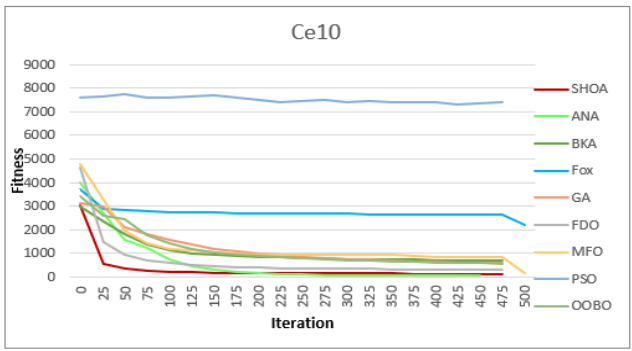
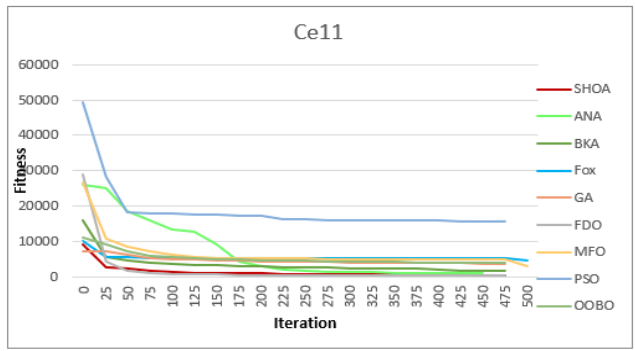
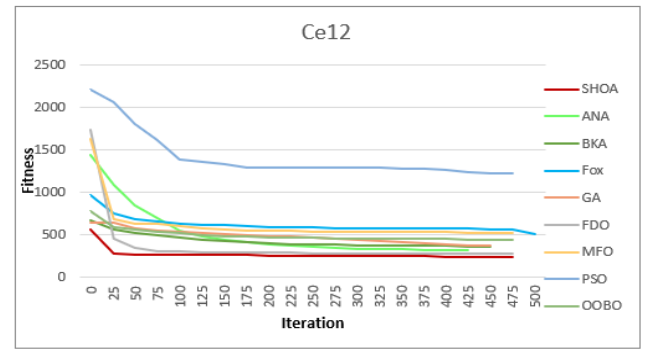

Figure 9 Convergence Curve CEC22 Test Instances

Table 12 Comparison of SHOA with winner CEC 2019

| Alg. | SHOA | | JDE100 | |
|---|---|---|---|---|
| F | Mean | Std | Mean | Std |
| C01 | 2.18E-01 | 2.18E-01 | 1.59E+05 | 1.597E+05 |
| C02 | 3.00E+00 | 3.00E+00 | 2.385E+06 | 2.719E+04 |
| C03 | 3.13E+00 | 3.13E+00 | 1.31E+06 | 8.519E+05 |
| C04 | 4.36E+01 | 4.36E+01 | 3.475E+05 | 1.149E+05 |
| C05 | 1.33E-01 | 1.33E-01 | 1.673E+05 | 8.426E+04 |
| C06 | 7.97E+00 | 7.97E+00 | 3.841E+04 | 2.063E+03 |
| C07 | 1.19E+02 | 1.19E+02 | 9.105E+06 | 4.528E+06 |
| C08 | 2.52E+00 | 2.52E+00 | 1.219E+09 | 4.388E+08 |
| C09 | 1.02E+00 | 1.02E+00 | 9.207E+08 | 1.131E+08 |
| C10 | 1.72E+00 | 1.72E+00 | 1.541E+06 | 7.46E+05 |

Table 13 Compare Values Of CEC 2022 for Different Dimension

| Alg. | D=2 | D=10 | D=20 |
|---|---|---|---|
| F | Mean | Mean | Mean |
| Ce1 | 1.09E-07 | 1.63E+00 | 1.84E+03 |
| Ce2 | 1.09E-07 | 2.45E-01 | 4.51E+02 |
| Ce3 | 1.07E-10 | 1.18E-05 | 5.39E-02 |
| Ce4 | 4.01E-05 | 1.21E+01 | 9.97E+01 |
| Ce5 | 4.18E-09 | 6.97E-03 | 1.87E+00 |
| Ce6 | - | 6.22E-01 | 7.13E+01 |
| Ce7 | - | 4.16E+00 | 2.74E+01 |
| Ce8 | - | 4.16E+00 | 2.39E+01 |
| Ce9 | 9.28E-03 | 1.87E+02 | 1.91E+02 |
| Ce10 | -1.14E+01 | 1.01E+02 | 1.01E+02 |
| Ce11 | 1.57E-02 | 1.69E+01 | 1.33E+02 |
| Ce12 | 8.91E-01 | 1.92E+02 | 2.26E+02 |

Table 14 Comparison results of CEC22 with different nestling for D=10

| Nest size | | B=4 | B=5 | B=5 | B=6 | B=7 |
|---|---|---|---|---|---|---|
| Ce1 | Mean | 3.06E+00 | 2.21E+00 | 2.06E+00 | | 1.63E+00 |
| | std | 1.70E+00 | 1.08E+00 | 9.97E-01 | | 6.54E-01 |
| Ce6 | Mean | 7.12E-01 | 9.08E-01 | 8.39E-01 | | 6.22E-01 |
| | std | 5.04E-01 | 5.13E-01 | 4.54E-01 | | 4.28E-01 |
| Ce12 | Mean | 1.94E+02 | 1.94E+02 | 1.93E+02 | | 1.92E+02 |
| | std | 1.46E+00 | 1.94E+02 | 1.45E+00 | | 1.30E+00 |

## IV. ENGINEERING PROBLEMS SOLVING

In this study, four constrained engineering problems, namely three-bar truss design, gear train design, antenna array design, and frequency-modulated sound wave design, are considered to investigate the applicability of SHOA. The problems have equality and inequality constraints, the SHOA should be equipped with the constrained solutions. Although, in constraint problem solving there will be feasible and infeasible solutions, to investigate infeasible solutions, some algorithms use penalty functions [96], in this study the death penalty is used, and the infeasible solutions are discarded and not investigated with a penalty to speed up the algorithm process. It is worth noting that the population size is set to 15, and iterations set to 500, for 30 rounds for all the problems in this section.

### A. GEAR TRAIN DESIGN PROBLEM SOLVING

The gear train design is a mechanical engineering problem, the main objective is to minimize the desired ratio with the current ratio [97], the objective function was formulated as follows:

$$f\left(\vec{x}\right) = \left(\frac{1}{6.931} - \frac{G_a G_b}{G_c G_d}\right)^2 \quad (11)$$

where $\frac{1}{6.931}$ desired ratio, Ga, Gb, Gc, Gd teeth of gears A, B, C, and D respectively, with the ratio is:

$$\text{Gear Ratio} = \frac{G_a G_b}{G_c G_d} \quad (12)$$

subject to: $\forall \{G_i, \ 12 \leq G_i \leq 60\}$, where $G_i$ is teeth of Ga, Gb, Gc, Gd.

Table 15 Comparative Result Gear train design problem

| Algorithm | x1 (Ga) | x2 (Gb) | x3 (Gc) | x4 (Gd) | Optimal Error | Ratio |
|---|---|---|---|---|---|---|
| SHOA | 27.47 | 16.60 | 57.04 | 55.41 | 3.21E-20 | 0.1442 |
| NL | 18 | 22 | 45 | 60 | 5.70E-04 | 0.1466 |
| CS | 19 | 16 | 43 | 49 | 2.70E-12 | 0.1442 |
| AZOA | 60 | 17.52 | 12 | 24.29 | 0 | 3.606 |
| MFO | 43 | 19 | 16 | 49 | 2.70E-12 | 1.042 |

In Table 15 for SHOA with AZOA [98], MFO, Non-Linear (NL) [97], and Cuckoo Search (CS) [99] shown, the table presents gear teeth of A, B, C, and D, optimal, and ratio (x) as comparison parameters, where the ratio must be closer to (1/6.931). In this study, SHOA had high performance over other algorithms, MFO has good optimal error but the (ratio > 1.442) CS also performed well in the second stage, but AZOA had bad performance because their ratio rates as constraints were not satisfied.

### B. THREE-BAR TRUSS DESIGN PROBLEM SOLVING

The three-bar truss problem is a civil engineering design problem whose objective is to achieve the minimum weight subjected to stress, deflection, and buckling constraints and evaluate the optimal cross-sectional area ($A1$, $A2$). Mathematically, to minimize the weight of a three-bar truss construction, according to [100], an objective function and constraints are formulated as follows:

Minimize $f(x) = (2\sqrt{2}x_1 + x_2) \times l$ (13)

Subject to:

$$C_1(x) = \frac{\sqrt{2} x_1 + x_2}{\sqrt{2} x^2_1 + 2x_1 x_2} P - \sigma \leq 0 \quad (14)$$

$$C_2(x) = \frac{x_2}{\sqrt{2} x^2_1 + 2x_1 x_2} P - \sigma \leq 0 \quad (15)$$

$$C_3(x) = \frac{1}{\sqrt{2} x_2 + x_1} P - \sigma \leq 0 \quad (16)$$

$\forall i, 0 \leq x_i \leq 1, where\ i = 1,2$, the constant parameters are: $l = 100\ cm$, $P = 2\ KN/cm^2$, $\sigma = 2\ KN/cm^2$.

The comparative result shown in Table 16 presents the performance of the SHOA compared with AZOA, CS, MFO, and engineering design optimization by (Ray T, and Saini) TSa [100] algorithms, SHOA had tested nearly 20 times in 30 rounds all minimum fitness was between (263.89, and 263.90) and maximum fitness was between (263.94, and 263.97).

Table 16 shows the performance of SHOA either better or equal other algorithms.

Table 16 Comparative Result Three-bar Truss Design Problem

| Algorithm | $x_1$ [$A_1$] | $x_2$ [$A_2$] | Optimal weight |
|---|---|---|---|
| SHOA | 0.788235 | 0.409503 | 263.8968 |
| TSa | 0.795 | 0.395 | 264.3 |
| CS | 0.788670 | 0.40902 | 263.9716 |
| AZOA | 0.7885471 | 0.408610 | 263.8958 |
| MFO | 0.7882447 | 0.4094669 | 263.8959 |

## C. ANTENNA SPACED ARRAY PROBLEM SOLVING WITH SHOA

Optimization of antenna arrays means reducing the side-lobe level (SLL) of a non-uniformly spaced linear array. The fitness value for the problem has been formulated to the maximum SLL to optimize the non-uniformly spaced array [25,101]. Objective function and constraints are formulated as follows:

$$f\left(\vec{x}\right) = max[20\, log\, |G(\theta)|] \qquad (17)$$

where the:
$$G(\theta) = \sum_{i=1}^{n} cos[2\pi x_i (cos\theta - cos\theta_s)] + cos[2.25 \times 2\pi(cos\theta - cos\theta_s)] \qquad (18)$$

$$n = 4 \quad and\ \theta = 45°, \quad \theta_b = 90°$$

Subject to:

$$d = |xi - xj| > 0.25\lambda \qquad (19)$$

$$0.125\lambda < min\{x_i\} \le 2.0\lambda \qquad (20)$$

$$x_i \in (0, 2.25),\ i = 1,2,3,4.\ i \ne j$$

To minimize SLL, the element should optimize without violation of above constraints above, where $\theta\ is\ elevation angle, and\ \theta_b\ is\ beam\ angle$, xi which is an element of the antenna must be greater than $0.125\lambda$, the distance between elements must be more than $0.25\lambda$.

Table 17 Comparative Result Antenna Spaced Array Problem

| Algorithm | $P_1$ | $P_2$ | $P_3$ | $P_4$ | Optimal SLL |
|---|---|---|---|---|---|
| SHOA | 0.584 | 1.288 | 1.865 | 1.599 | -280.491 |
| FDO | 0.713 | 1.595 | 0.433 | 0.130 | -120 |
| ANA | 1.5959 | 0.3081 | 0.8747 | 0.6072 | -70.720 |

The comparative assessment in Table 17 shows an optimal value between SHOA and other algorithms that applied antenna array space problem, the SHOA found a minimum optimal out of 30 rounds as shown in Table 17, and the maximum optimal value was (-177.46) with parameters (1.237, 0.789, 1.513, 0.432), the assessment shows the superiority of SHOA in all rounds when compared with FDO and ANA.

## D. FREQUENCY-MODULATED SOUND WAVE DESIGN PROBLEM SOLVING

The frequency modulation in sound waves is required to find optimal parameters to transfer sounds, it has six parameters to optimize as ($a1, w1, a2, w2, a3, w3$), which is a highly complex problem in the multimodal field, with fitness function is a minimum summation of square error between evaluated and modeled data, the fitness and constraints are formulated as follows:

$$f\left(\vec{p}\right) = \sum_{i=1}^{100}(y(t) - y_0(t))^2 \qquad (21)$$

Where:
$$\left(\vec{p}\right) = (a1, w1, a2, w2, a3, w3) \qquad (22)$$

$$y(t) = a_1.sin(w_1.t.\theta + a_2.sin(w_2.t.\theta + a_3.sin(w_3.t.\theta))) \qquad (23)$$

$$y(t) = (1.0).sin((5.0).t.\theta + (1.5).sin((4.8).t.\theta + (2.0).sin((4.9).t.\theta))) \qquad (24)$$

With $\theta = (2\pi/100)$, the range of the parameter is [ −6.4, 6.35], and minimum fitness values are the optimal solution for the sound wave problems to transfer sound with the lowest error rate [94,102,103].

Table 18 Comparative result of frequency-modulated sound wave

| Alg. | $a_1$ | $w_1$ | $a_2$ | $w_2$ | $a_3$ | $w_3$ | Error | avg. |
|---|---|---|---|---|---|---|---|---|
| SHOA | 1.032 | 5.013 | 1.893 | 4.822 | 1.80 | 4.887 | 2.498 | 10.3 |
| FDO | 0.97 | -0.24 | -4.31 | -0.01 | -0.57 | 4.93 | 3.22 | NA |
| ANA | −0.12 | 3.05 | −5.21 | −4.44 | 2.56 | 2.85 | 53.249 | NA |
| fGA | NA | NA | NA | NA | NA | NA | 0.0 | 8.4 |

In Table 18 comparative results of a frequency-modulated wave for the SHOA and FDO, ANA and Fork Genetic Algorithm (fGA) [94] are shown, the table presented six parameters, with the best fitness value out of 30 runs considered as an optimal result, and an average of 30 runs of the SHOA with the mentioned algorithms FDO and fGA, the unknown data is written as NA. The result shows that SHOA has a higher performance than all algorithms. The SHOA finds better fitness than FDO and a better average than fGA out of 30 runs. The FDO average result was NA, an optimal value has been generated depending on the presented parameters from FDO, and the fGA reached the optimal solution $\left(\vec{p}\right) = \mathbf{0.0}$, but the parameters had not been presented.

## V. CONCLUSIONS

In this study, the theoretical oft for the novel SHO swarm-based algorithm has been provided via concepts of exploration and exploitation. It mimics the bird's breed adaptation and lifestyle in the population.

The SHOA applied to 41 benchmark functions (uni-modal, multi-modal, composite, and 100-Digit Challenge) test functions, CEC22 single objective bounded constrained optimization and engineering problems (constrained, unconstrained) are solved. The performance has been compared with recent and powerful algorithms. The results demonstrated the effectiveness of the newly developed approach SHOA in solving all test functions and a variety of engineering problems and showed that this can provide reliable and accurate solutions in a variety of contexts. Through the SHOA study, the following were concluded:

- Faster convergence rate, the mechanism adapts the best birds for the next generation.
- More stable than compared algorithms, the balance of convergence and divergence leads to the best solution.
- Accurate search, high exploration, and investigation promise a promising area of space within a reasonable amount of time.
- High performance in solving constrained and un-constrained multimodal real optimization problems.
- Highly multi-modal problem optimizer, because each nest is considered a population with an optimal solution, and all are considered a single population that finds the global optimum from local optimums.
- There are fewer control parameters compared with other optimization algorithms, which leads to low computation time.

The proposed SHOA is a single objective; for the future, many research works can be conducted in multi-objective, binary, and discrete versions, all of which can be used to solve a variety of types of problems.

**ACKNOWLEDGMENT**

The authors would like to sincerely thank the members of the software engineering department from Slahaddin University-Erbil for their invaluable contributions that enabled this study to be completed successfully.

## Appendix A

*Table 19* Uni-modal Test Functions with dimension = 10

| Function | Range | Shift | $f_{min}$ |
|---|---|---|---|
| $F1(x) = \sum_{i=1}^{n} x_i^2$ | [-100, 100] | [-30, -30, -30, …] | 0 |
| $F2(x) = \sum_{i=1}^{n}\|x_i\| + \prod_{i=1}^{n}\|x_i\|$ | [-10, 10] | [-3, -3, -3, …] | 0 |
| $F3(x) = \sum_{i=1}^{n}\left(\sum_{j=1}^{i} x_j\right)^2$ | [-100, 100] | [-30, -30, -30, …] | 0 |
| $F4(x) = max_i\{\|x_i\|, 1 \leq i \leq n\}$ | [-100, 100] | [-30, -30, -30, …] | 0 |
| $F5(x) = \sum_{i=1}^{n-1}[100(x_{i+1} - x_i^2)^2 + (x_i - 1)^2]$ | [-30, 30] | [-15, -15, -15, …] | 0 |
| $F6(x) = \sum_{i=1}^{n}(\lfloor x_i + 0.5 \rfloor)^2$ | [-100, 100] | [-75, -75, -75, …] | 0 |
| $F7(x) = \sum_{i=1}^{n} ix_i^4 + rand[0,1]$ | [-1.28, 1.28] | [-0.25, -0.25, -0.25, …] | 0 |

Table 20 Multi-modal Test Functions with dimension = 10

| Function | Range | Shift | min |
|---|---|---|---|
| $F8(x) = \sum_{i=1}^{n} -x_i^2 \sin(\sqrt{\|x_i\|})$ | [-500, 500] | [-300, -300, -300, …] | -418.9829 |
| $F9(x) = \sum_{i=1}^{n}[x_i^2 - \cos(2\pi x_i) + 10]$ | [-5.12, 5.12] | [-2, -2, -2, …] | 0 |
| $F10(x) = -20\,exp\left(-0.2\sqrt{\frac{1}{n}\sum_{i=1}^{n} x_i^2}\right) - exp\left(\frac{1}{n}\sum_{i=1}^{n} \cos(2\pi x_i)\right) + 20 + e$ | [-32, 32] | | 0 |
| $F11(x) = \frac{1}{4000}\sum_{i=1}^{n} x_i^2 - \prod_{i=1}^{n} \cos\left(\frac{x_i}{\sqrt{i}}\right) + 1$ | [-600, 600] | [-400, -400, -400, …] | 0 |
| $F12(x) = \frac{\pi}{n}\left\{10\sin^2(\pi y_1) + \sum_{i=1}^{n-1}(y_i - 1)^2[1 + 10\sin^2(\pi y_{i+1})] + (y_n - 1)^2\right\} + \sum_{i=1}^{n} u(x_i, 10, 100, 4)$ $y_i = 1 + \frac{(x+1)}{4}$ $u(x_i, a, k, m) = \begin{cases} k(x_i - a)^m & x_i > a \\ 0 & -a < x_i < a \\ k(-x_i - a)^m & x_i < -a \end{cases}$ | [-50, 50] | [-30, -30, -30, …] | 0 |
| $F13(x) = 0.1\left\{\sin^2(3\pi x_1) + \sum_{i=1}^{n-1}(x_i - 1)^2[1 + \sin^2(3\pi x_i + 1)] + (x_n - 1)[1 + \sin^2(2\pi x_n)]\right\} + \sum_{i=1}^{n} u(x_i, 5, 100, 4)$ $u(x_i, a, k, m) = \begin{cases} k(x_i - a)^m & x_i > a \\ 0 & -a < x_i < a \\ k(-x_i - a)^m & x_i < -a \end{cases}$ | [-50, 50] | [-10, -10, -10, …] | 0 |

Table 21 Composite Test Functions with dimension = 10, Range [-5,5], $f_{min} = 0$

| Functions |
|---|
| **F14 (CF1)** |
| $f1, f2, \ldots, f10 = $ sphere function |
| $[\delta_1, \delta_2, \delta_3 \ldots \delta_{10}] = [1,1,1,\ldots,1]$ |
| $[\lambda_1, \lambda_2, \ldots, \lambda_{10}] = \left[\frac{5}{100}, \frac{5}{100}, \ldots, \frac{5}{100}\right]$ |
| **F15 (CF2)** |
| $f1, f2, \ldots, f10 = $ Griewank's function |
| $[\delta_1, \delta_2, \delta_3 \ldots \delta_{10}] = [1,1,1,\ldots,1]$ |
| $[\lambda_1, \lambda_2, \ldots, \lambda_{10}] = \left[\frac{5}{100}, \frac{5}{100}, \ldots, \frac{5}{100}\right]$ |
| **F16 (CF3)** |
| $f1, f2, \ldots, f10 = $ Griewank's function |
| $[\delta_1, \delta_2, \delta_3 \ldots \delta_{10}] = [1,1,1,\ldots,1]$ |
| $[\lambda_1, \lambda_2, \ldots, \lambda_{10}] = [1,1,\ldots,1]$ |
| **F17 (CF4)** |
| $f1, f2 = $ Ackley's function |
| $f3, f4 = $ Rastrigin's function |
| $f5, f6 = $ Weierdtrass function |
| $f7, f8 = $ Griewank's function |
| $f9, f10 = $ Sphere function |
| $[\delta_1, \delta_2, \delta_3 \ldots \delta_{10}] = [1,1,1,\ldots,1]$ |
| $[\lambda_1, \lambda_2, \ldots, \lambda_{10}] = \left[\frac{5}{32}, \frac{5}{32}, 1, 1, \frac{5}{0.5}, \frac{5}{0.5}, \frac{5}{100}, \frac{5}{100}, \frac{5}{100}, \frac{5}{100}\right]$ |
| **F18 (CF4)** |
| $f1, f2 = $ Rastrigin's function |
| $f3, f4 = $ Weierdtrass function |
| $f5, f6 = $ Griewank's function |
| $f7, f8 = $ Ackley's function |
| $f9, f10 = $ Sphere function |
| $[\delta_1, \delta_2, \delta_3 \ldots \delta_{10}] = [1,1,1,\ldots,1]$ |
| $[\lambda_1, \lambda_2, \ldots, \lambda_{10}] = \left[\frac{1}{5}, \frac{1}{5}, \frac{5}{0.5}, \frac{5}{0.5}, \frac{5}{100}, \frac{5}{100}, \frac{5}{32}, \frac{5}{32}, \frac{5}{100}, \frac{5}{100}\right]$ |
| **F18 (CF4)** |
| $f1, f2 = $ Rastrigin's function |
| $f3, f4 = $ Weierdtrass function |
| $f5, f6 = $ Griewank's function |
| $f7, f8 = $ Ackley's function |
| $f9, f10 = $ Sphere function |
| $[\delta_1, \delta_2, \delta_3 \ldots \delta_{10}] = [0.1, 0.2, 0.3, 0.4, 0.5, 0.6, 0.7, 0.9, 1]$ |
| $[\lambda_1, \lambda_2, \ldots, \lambda_{10}] = \left[0.1 \times \frac{1}{5}, 0.2 \times \frac{1}{5}, 0.3 \times \frac{5}{0.5}, 0.4 \times \frac{5}{0.5}, 0.5 \times \frac{5}{100}, 0.6 \times \frac{5}{100}, 0.7 \times \frac{5}{32}, 0.8 \times \frac{5}{32}, 0.9 \times \frac{5}{100}, 1 \times \frac{5}{100}\right]$ |

Table 22 Summary of basic "The Hundred-Digit Challenge" Benchmarks

| Name | Functions | Dim | Range |
|---|---|---|---|
| **C01** | STRONG CHEBYSHEV POLYNOMIAL FITTING PROBLEM | 9 | [-8192,8192] |
| **C02** | INVERSE HILBERT MATRIX PROBLEM | 16 | [-16384, 16384] |
| **C03** | LENNARD-JONES MINIMUM ENERGY CLUSTER | 18 | [-4,4] |
| **C04** | RASTRIGIN'S FUNCTION | 10 | [-100, 100] |
| **C05** | GRIEWANGK'S FUNCTION | 10 | [-100, 100] |
| **C06** | WEIERSTRASS FUNCTION | 10 | [-100, 100] |
| **C0** | MODIFIED SCHWEFEL'S FUNCTION | 10 | [-100, 100] |
| **C08** | EXPANDED SCHAFFER'S F6 FUNCTION | 10 | [-100, 100] |
| **C09** | HAPPY CAT FUNCTION | 10 | [-100, 100] |
| **C10** | ACKLEY FUNCTION | 10 | [-100, 100] |

Table 23 Summary of CEC2022 benchmarks

| Types | No | Functions | $F_i^*$ |
|---|---|---|---|
| Unimodal | 1 | Shifted and full Rotated Zakharov Function | 300 |
| Basic Functions | 2 | Shifted and full Rotated Rosenbrock's Function | 400 |
| | 3 | Shifted and full Rotated Expanded Schaffer's f6 Function | 600 |
| | 4 | Shifted and full Rotated Non-Continuous Rastrigin's Function | 800 |
| | 5 | Shifted and full Rotated Levy Function | 900 |
| Hybrid multimodal Functions | 6 | Hybrid Function 1 (N = 3) | 1800 |
| | 7 | Hybrid Function 2 (N = 6) | 2000 |
| | 8 | Hybrid Function 3 (N = 5) | 2200 |
| Composite multimodal Functions | 9 | Composition Function 1 (N = 5) | 2300 |
| | 10 | Composition Function 2 (N = 4) | 2400 |
| | 11 | Composition Function 3 (N = 5) | 2600 |
| | 12 | Composition Function 4 (N = 6) | 2700 |
| Search Range [-100, 100] | | | |

**REFERENCES**

[1] Kirkpatrick S, Gelatt CD, Vecchi MP. Optimization by Simulated Annealing. Science (80- ) 1983;220:4598. https://doi.org/http://science.sciencemag.org/.

[2] Greiner R. PALO: A probabilistic hill-climbing algorithm. Artif Intell 1996;84:177–208. https://doi.org/10.1016/0004-3702(95)00040-2.

[3] Hansen P, Mladenović N. Variable neighborhood search. Handb Heuristics 1997;1–2:1097–100. https://doi.org/10.1007/978-3-319-07124-4_19.

[4] Glover F. Tabu Search: A Tutorial. Interfaces (Providence) 1990;20:74–94. https://doi.org/10.1287/inte.20.4.74.

[5] Holland JH 1975. Adaptation in Natural and Artificial Systems:An introductory analysis with applications to biology, control, and artificial intelligence. 2nd Editio. The MIT Press; n.d. https://doi.org/https://psycnet.apa.org/record/1975-26618-000.

[6] RAINER Storn KP. Differential Evolution – A Simple and Efficient Heuristic for Global Optimization over Continuous Spaces. Australas Plant Pathol 1997;38:284–7. https://doi.org/10.1071/AP09004.

[7] Koza. JR. Hierarchical Genetic Algorithms Operating on Populations of Computer Programs. Hierarchical Genet Algorithms Oper Popul Comput Programs 1989;1:768–774.

[8] Dorigo, M., Birattari, M., & Stutzle T. Ant Colony Optimization. IEEE Comput Intell Mag 2006;1(4):28–39. https://doi.org/doi:10.1109/mci.2006.329691.

[9] Dorigo M, Maniezzo V, Colorni A. Ant system: Optimization by a colony of cooperating agents. IEEE Trans Syst Man, Cybern Part B Cybern 1996;26:29–41. https://doi.org/10.1109/3477.484436.

[10] Dorigo M. Learning and Natural Algorithms 1992.
[11] Martens D, De Backer M, Haesen R, Vanthienen J, Snoeck M, Baesens B. Classification with ant colony optimization. IEEE Trans Evol Comput 2007;11:651–65. https://doi.org/10.1109/TEVC.2006.890229.
[12] Zuo L, Shu L, Dong S, Zhu C, Hara T. A multi-objective optimization scheduling method based on the ant colony algorithm in cloud computing. IEEE Access 2015;3:2687–99. https://doi.org/10.1109/ACCESS.2015.2508940.
[13] Rashid DNH, Rashid TA, Mirjalili S. Ana: Ant nesting algorithm for optimizing real-world problems. Mathematics 2021;9:1–30. https://doi.org/10.3390/math9233111.
[14] Eberhart R, Kennedy J. New optimizer using particle swarm theory. Proc Int Symp Micro Mach Hum Sci 1995:39–43. https://doi.org/10.1109/mhs.1995.494215.
[15] Braik M, Sheta A, Ayesh A. Image Enhancement Using Particle Swarm Optimization. Int J Innov Comput Appl 2007;1:138–45.
[16] Mikki SM, Kishk AA. Quantum particle swarm optimization for electromagnetics. IEEE Trans Antennas Propag 2006;54:2764–75. https://doi.org/10.1109/TAP.2006.882165.
[17] Ray T, Liew KM. Society and civilization: An optimization algorithm based on the simulation of social behavior. IEEE Trans Evol Comput 2003;7:386–96. https://doi.org/10.1109/TEVC.2003.814902.
[18] Karaboga D. An idea based on honey bee swarm for numerical optimization. Tech ReportTR06, Erciyes Univ Eng Fac Comput Eng Dep 2005.
[19] Karaboga D, Basturk B. A powerful and efficient algorithm for numerical function optimization: Artificial bee colony (ABC) algorithm. J Glob Optim 2007;39:459–71. https://doi.org/10.1007/s10898-007-9149-x.
[20] Gao W, Liu S, Huang L. A global best artificial bee colony algorithm for global optimization. J Comput Appl Math 2012;236:2741–53. https://doi.org/10.1016/j.cam.2012.01.013.
[21] Uzer MS, Yilmaz N, Inan O. Feature selection method based on artificial bee colony algorithm and support vector machines for medical datasets classification. Sci World J 2013;2013. https://doi.org/10.1155/2013/419187.
[22] Bullinaria JA, AlYahya K. Artificial Bee Colony training of neural networks: Comparison with back-propagation. Memetic Comput 2014;6:171–82. https://doi.org/10.1007/s12293-014-0137-7.
[23] Yao B, Yan Q, Zhang M, Yang Y. Improved artificial bee colony algorithm for vehicle routing problem with time windows. PLoS One 2017;12:1–18. https://doi.org/10.1371/journal.pone.0181275.
[24] El-abd M. Generalized Opposition-Based Artificial Bee Colony Algorithm 2012.
[25] Abdullah JM, Ahmed T. Fitness Dependent Optimizer: Inspired by the Bee Swarming Reproductive Process. IEEE Access 2019;7:43473–86. https://doi.org/10.1109/ACCESS.2019.2907012.
[26] Passino KM. Bacterial Foraging Optimization. Int J Swarm Intell Res 2010;1:1–16. https://doi.org/10.4018/jsir.2010010101.
[27] Patnaik SS, Panda AK. Particle Swarm Optimization and Bacterial Foraging Optimization Techniques for Optimal Current Harmonic Mitigation by Employing Active Power Filter. Appl Comput Intell Soft Comput 2012;2012:1–10. https://doi.org/10.1155/2012/897127.
[28] Muñoz MA, Halgamuge SK, Alfonso W, Caicedo EF. Simplifying the bacteria foraging optimization algorithm. 2010 IEEE World Congr Comput Intell WCCI 2010 - 2010 IEEE Congr Evol Comput CEC 2010 2010:18–23. https://doi.org/10.1109/CEC.2010.5586025.
[29] Yang XS. Firefly algorithms for multimodal optimization. Lect Notes Comput Sci (including Subser Lect Notes Artif Intell Lect Notes Bioinformatics) 2009;5792 LNCS:169–78. https://doi.org/10.1007/978-3-642-04944-6_14.
[30] Gebremedhen HS, Woldemichael DE, Hashim FM. A firefly algorithm based hybrid method for structural topology optimization. Adv Model Simul Eng Sci 2020;7. https://doi.org/10.1186/s40323-020-00183-0.
[31] Mirjalili S. Moth-flame optimization algorithm: A novel nature-inspired heuristic paradigm. Knowledge-Based Syst 2015;89:228–49. https://doi.org/10.1016/j.knosys.2015.07.006.
[32] Pedram Haeri Boroujeni S, Pashaei E. Data Clustering Using Moth-F lame Optimization Algorithm. ICCKE 2021 - 11th Int Conf Comput Eng Knowl 2021:296–301. https://doi.org/10.1109/ICCKE54056.2021.9721483.
[33] Wang X, Snášel V, Mirjalili S, Pan JS, Kong L, Shehadeh HA. Artificial Protozoa Optimizer (APO): A novel bio-inspired metaheuristic algorithm for engineering optimization. Knowledge-Based Syst 2024;295. https://doi.org/10.1016/j.knosys.2024.111737.
[34] Peraza-Vázquez H, Peña-Delgado A, Merino-Treviño M, Morales-Cepeda AB, Sinha N. A novel metaheuristic inspired by horned lizard defense tactics. vol. 57. Springer Netherlands; 2024. https://doi.org/10.1007/s10462-023-10653-7.
[35] Wang J, Wang WC, Hu XX, Qiu L, Zang HF. Black-winged kite algorithm: a nature-inspired meta-heuristic for solving benchmark functions and engineering problems. vol. 57. Springer

Netherlands; 2024. https://doi.org/10.1007/s10462-024-10723-4.

[36] Oladejo SO, Ekwe SO, Mirjalili S. The Hiking Optimization Algorithm: A novel human-based metaheuristic approach. Knowledge-Based Syst 2024;296:111880. https://doi.org/10.1016/j.knosys.2024.111880.

[37] Baskar A, Xavior MA, Jeyapandiarajan P, Batako A, Burduk A. A novel two-phase trigonometric algorithm for solving global optimization problems. Ann Oper Res 2024. https://doi.org/10.1007/s10479-024-05837-5.

[38] Osaba E, Yang X-S. Soccer-Inspired Metaheuristics: Systematic Review of Recent Research and Applications 2021:81–102. https://doi.org/10.1007/978-981-16-0662-5_5.

[39] Jain M, Singh V, Rani A. A novel nature-inspired algorithm for optimization: Squirrel search algorithm. Swarm Evol Comput 2019;44:148–75. https://doi.org/10.1016/j.swevo.2018.02.013.

[40] Mugemanyi S, Qu Z, Rugema FX, Dong Y, Wang L, Bananeza C, et al. Marine predators algorithm: A comprehensive review. Mach Learn with Appl 2023;12:100471. https://doi.org/10.1016/j.mlwa.2023.100471.

[41] Chaudhary, Reshu banati. PeacockAlgorithm 2019.

[42] Akbari MA, Zare M, Azizipanah-abarghooee R, Mirjalili S, Deriche M. The cheetah optimizer: a nature-inspired metaheuristic algorithm for large-scale optimization problems. Sci Rep 2022;12:1–20. https://doi.org/10.1038/s41598-022-14338-z.

[43] Abdollahzadeh B, Gharehchopogh FS, Khodadadi N, Mirjalili S. Mountain Gazelle Optimizer: A new Nature-inspired Metaheuristic Algorithm for Global Optimization Problems. Adv Eng Softw 2022;174:103282. https://doi.org/10.1016/j.advengsoft.2022.103282.

[44] Wan Afandie WNEA, Rahman TKA, Zakaria Z. Optimal load shedding using Bacterial Foraging Optimization Algorithm. Proc - 2013 IEEE 4th Control Syst Grad Res Colloquium, ICSGRC 2013 2013:93–7. https://doi.org/10.1109/ICSGRC.2013.6653282.

[45] Mangaraj BB, Jena MR, Mohanty SK. Bacteria Foraging Algorithm in Antenna Design. Appl Comput Intell Soft Comput 2016;2016. https://doi.org/10.1155/2016/5983469.

[46] Skackauskas J, Kalganova T, Dear I, Janakiram M. Dynamic impact for ant colony optimization algorithm. Swarm Evol Comput 2022;69. https://doi.org/10.1016/j.swevo.2021.100993.

[47] Dokeroglu T, Sevinc E, Cosar A. Artificial bee colony optimization for the quadratic assignment problem. Appl Soft Comput J 2019;76:595–606. https://doi.org/10.1016/j.asoc.2019.01.001.

[48] Z. Xu CF. Design of Available Land Space Optimization Model Based on Ant Colony Optimization Algorithm. IEEE Int. Conf. Integr. Circuits Commun. Syst., Raichur, India: 2023, bl 1–6. https://doi.org/10.1109/ICICACS57338.2023.10099535.

[49] Qolomany B, Ahmad K, Al-Fuqaha A, Qadir J. Particle Swarm Optimized Federated Learning for Industrial IoT and Smart City Services. 2020 IEEE Glob Commun Conf GLOBECOM 2020 - Proc 2020. https://doi.org/10.1109/GLOBECOM42002.2020.9322464.

[50] Zou R, Kalivarapu V, Winer E, Oliver J, Bhattacharya S. Particle Swarm Optimization-Based Source Seeking. IEEE Trans Autom Sci Eng 2015;12:865–75. https://doi.org/10.1109/TASE.2015.2441746.

[51] Djemai T, Stolf P, Monteil T, Pierson JM. A discrete particle swarm optimization approach for energy-efficient IoT services placement over fog infrastructures. Proc - 2019 18th Int Symp Parallel Distrib Comput ISPDC 2019 2019;2019:32–40. https://doi.org/10.1109/ISPDC.2019.00020.

[52] Abdulkarim HK, Rashid TA. Moth-Flame Optimization and Ant Nesting Algorithm : A Systematic Evaluation. Proc. 1st Int. Conf. Innov. Inf. Technol. Bus. (ICIITB 2022), vol. 1, Online: Atlantis Press; 2023, bl 139–52. https://doi.org/10.2991/978-94-6463-110-4_11.

[53] Salim A, Jummar WK, Jasim FM, Yousif M. Eurasian oystercatcher optimiser: New meta-heuristic algorithm. J Intell Syst 2022;31:332–44. https://doi.org/10.1515/jisys-2022-0017.

[54] Braik M, Hammouri A, Atwan J, Al-Betar MA, Awadallah MA. White Shark Optimizer: A novel bio-inspired meta-heuristic algorithm for global optimization problems. Knowledge-Based Syst 2022;243:108457. https://doi.org/https://doi.org/10.1016/j.knosys.2022.108457.

[55] Mohammed H. FOX : a FOX-inspired optimization algorithm. Appl Intell 2023;53:1030–1050.

[56] Golilarz NA, Gao H, Addeh A, Pirasteh S. ORCA Optimization Algorithm: A New Meta-Heuristic Tool for Complex Optimization Problems. 2020 17th Int Comput Conf Wavelet Act Media Technol Inf Process ICCWAMTIP 2020 2020:198–204. https://doi.org/10.1109/ICCWAMTIP51612.2020.9317473.

[57] Trojovský P, Dehghani M. A new bio-inspired metaheuristic algorithm for solving optimization problems based on walruses behavior. vol. 13. Nature Publishing Group UK; 2023. https://doi.org/10.1038/s41598-023-35863-5.

[58] Vela J, Montiel EE, Mora P, Lorite P, Palomeque T. Aphids and ants, mutualistic species, share a Mariner element with an unusual location on aphid

chromosomes. Genes (Basel) 2021;12. https://doi.org/10.3390/genes12121966.
[59] Azizi M, Talatahari S, Gandomi AH. Fire Hawk Optimizer: a novel metaheuristic algorithm. vol. 56. Springer Netherlands; 2023. https://doi.org/10.1007/s10462-022-10173-w.
[60] Hashim FA, Houssein EH, Hussain K, Mabrouk MS, Al-Atabany W. Honey Badger Algorithm: New metaheuristic algorithm for solving optimization problems. Math Comput Simul 2022;192:84–110. https://doi.org/10.1016/j.matcom.2021.08.013.
[61] Kaur S, Awasthi LK, Sangal AL, Dhiman G. Tunicate Swarm Algorithm: A new bio-inspired based metaheuristic paradigm for global optimization. Eng Appl Artif Intell 2020;90:103541. https://doi.org/10.1016/j.engappai.2020.103541.
[62] Al-Baik O, Alomari S, Alssayed O, Gochhait S, Leonova I, Dutta U, et al. Pufferfish Optimization Algorithm: A New Bio-Inspired Metaheuristic Algorithm for Solving Optimization Problems. Biomimetics 2024;9. https://doi.org/10.3390/biomimetics9020065.
[63] Abdelhamid AA, Towfek SK, Khodadadi N, Alhussan AA, Khafaga DS, Eid MM, et al. Waterwheel Plant Algorithm: A Novel Metaheuristic Optimization Method. Processes 2023;11:1–25. https://doi.org/10.3390/pr11051502.
[64] Kaveh A, Seddighian MR, Ghanadpour E. Black Hole Mechanics Optimization: a novel meta-heuristic algorithm. Asian J Civ Eng 2020;21:1129–49. https://doi.org/10.1007/s42107-020-00282-8.
[65] Trojovská E, Dehghani M. A new human-based metahurestic optimization method based on mimicking cooking training. Sci Rep 2022;12:1–25. https://doi.org/10.1038/s41598-022-19313-2.
[66] Said M, Houssein EH, Deb S, Alhussan AA, Ghoniem RM. A Novel Gradient Based Optimizer for Solving Unit Commitment Problem. IEEE Access 2022;10:18081–92. https://doi.org/10.1109/ACCESS.2022.3150857.
[67] Dehghani M, Trojovská E, Trojovský P, Malik OP. OOBO: A New Metaheuristic Algorithm for Solving Optimization Problems. Biomimetics 2023;8. https://doi.org/10.3390/biomimetics8060468.
[68] Gao Y. PID-based search algorithm: A novel metaheuristic algorithm based on PID algorithm. Expert Syst Appl 2023;232:120886. https://doi.org/10.1016/j.eswa.2023.120886.
[69] Braik M, Ryalat MH, Al-Zoubi H. A novel meta-heuristic algorithm for solving numerical optimization problems: Ali Baba and the forty thieves. vol. 34. Springer London; 2022. https://doi.org/10.1007/s00521-021-06392-x.
[70] Jiang Y, Zhan ZH, Tan KC, Zhang J. Optimizing Niche Center for Multimodal Optimization Problems. IEEE Trans Cybern 2023;53:2544–57. https://doi.org/10.1109/TCYB.2021.3125362.
[71] Gálvez J, Cuevas E, Dhal KG. A competitive memory paradigm for multimodal optimization driven by clustering and chaos. Mathematics 2020;8. https://doi.org/10.3390/math8060934.
[72] Poole DJ, Allen CB. Constrained niching using differential evolution. Swarm Evol Comput 2019;44:74–100. https://doi.org/10.1016/j.swevo.2018.11.004.
[73] Agrawal S, Tiwari A, Naik P, Srivastava A. Improved differential evolution based on multi-armed bandit for multimodal optimization problems. Appl Intell 2021;51:7625–46. https://doi.org/10.1007/s10489-021-02261-1.
[74] Chen ZG, Zhan ZH, Wang H, Zhang J. Distributed Individuals for Multiple Peaks: A Novel Differential Evolution for Multimodal Optimization Problems. IEEE Trans Evol Comput 2020;24:708–19. https://doi.org/10.1109/TEVC.2019.2944180.
[75] Yan X, Razeghi-Jahromi M, Homaifar A, Erol BA, Girma A, Tunstel E. A novel streaming data clustering algorithm based on fitness proportionate sharing. IEEE Access 2019;7:184985–5000. https://doi.org/10.1109/ACCESS.2019.2922162.
[76] Joy G, Huyck C, Yang X-S. Review of Parameter Tuning Methods for Nature-Inspired Algorithms 2023;47:33–47. https://doi.org/10.1007/978-981-99-3970-1_3.
[77] Chaudhary R, Banati H. Study of population partitioning techniques on efficiency of swarm algorithms. Swarm Evol Comput 2020;55:100672. https://doi.org/10.1016/j.swevo.2020.100672.
[78] Li Q, Liu SY, Yang XS. Influence of initialization on the performance of metaheuristic optimizers. Appl Soft Comput J 2020;91:106193. https://doi.org/10.1016/j.asoc.2020.106193.
[79] Yazdani D, Omidvar MN, Yazdani D, Branke J, Nguyen TT, Gandomi AH, et al. Robust Optimization Over Time: A Critical Review. IEEE Trans Evol Comput 2023:1–21. https://doi.org/10.1109/TEVC.2023.3306017.
[80] Salgotra R, Sharma P, Raju S, gandomi AH. A Contemporary Systematic Review on Meta-heuristic Optimization Algorithms with Their MATLAB and Python Code Reference. vol. 31. Springer Netherlands; 2024. https://doi.org/10.1007/s11831-023-10030-1.
[81] Chaudhary R, Banati H. Capitalizing Diversity for Efficiency Enhancement in Multi-Population Swarm Algorithms. 2019 10th Int Conf Comput Commun Netw Technol ICCCNT 2019 2019:1–7. https://doi.org/10.1109/ICCCNT45670.2019.8944872.

[82] Chaudhary R, Banati H. Improving convergence in swarm algorithms by controlling range of random movement. vol. 20. Springer Netherlands; 2021. https://doi.org/10.1007/s11047-020-09826-y.

[83] Hill C, Miles K, Maddox K, Tegeler A. y to aid mate acquisition?Are adoptions in a predatory songbird a strateg. J F Ornithol 2023;94:0–5. https://doi.org/10.5751/jfo-00284-940209.

[84] Yosef R, Grubb TC. Territory Size Influences Nutritional Condition in Nonbreeding Loggerhead Shrikes ( Lanius ludovicianus ): A Ptilochronology Approach . Conserv Biol 1992;6:447–9. https://doi.org/10.1046/j.1523-1739.1992.06030447.x.

[85] Gawlik DE, Bildstein KL. Reproductive success and nesting habitat of Loggerhead Shrikes in north-central South Carolina. Wilson Bull 1990;102:37–48. https://doi.org/http://www.ncbi.nlm.nih.gov/pubmed/24671971.

[86] Ridgway R. A manual of north american birds. PHILADEPHIA: J.B. LIPPNCOTT COMPANY: Contributed by University of Toronto - Gerstein Science Information Centre.; 1887. https://doi.org/https://www.biodiversitylibrary.org/page/21816472.

[87] Yao X, Liu Y, Lin G. Evolutionary programming made faster. IEEE Trans Evol Comput 1999;3:82–102. https://doi.org/10.1109/4235.771163.

[88] Liang JJ, Suganthan PN, Deb K. Novel composition test functions for numerical global optimization. Proc - 2005 IEEE Swarm Intell Symp SIS 2005 2005;2005:71–8. https://doi.org/10.1109/SIS.2005.1501604.

[89] Marcin Molga CS. Test Functions for optimization algorithm 2005:205–9.

[90] Derrac J, García S, Molina D, Herrera F. A practical tutorial on the use of nonparametric statistical tests as a methodology for comparing evolutionary and swarm intelligence algorithms. Swarm Evol Comput 2011;1:3–18. https://doi.org/10.1016/j.swevo.2011.02.002.

[91] Iverson BL, Dervan PB. Problem Definitions and Evaluation Criteria for the CEC 2005 Special Session on Real-Parameter Optimization n.d.:7823–30.

[92] K. V. Price, N. H. Awad, M. Z. Ali PNS. Problem Definitions and Evaluation Criteria for the 100-Digit Challenge Special Session and Competition on Single Objective Numerical Optimization. Tech Rep Nanyang Technol Univ Singapore 2018:22.

[93] Kumar A, Price K V, Mohamed AW, Hadi AA, Suganthan PN. Problem Definitions and Evaluation Criteria for the CEC 2022 Special Session and Competition on Single Objective Bound Constrained Numerical Optimization. 2022 IEEE Congr Evol Comput 2022:1–20.

[94] Tsutsui S, Fujimoto Y. Forking Genetic Algorithm with Blocking and Shrinking Modes (fGA). Proc 5th Int Conf Genet Algorithms, ICGA '93 1993:206–15.

[95] Brest J, Maucec MS, Boskovic B. The 100-Digit Challenge: Algorithm jDE100. 2019 IEEE Congr Evol Comput CEC 2019 - Proc 2019:19–26. https://doi.org/10.1109/CEC.2019.8789904.

[96] Coello Coello CA. Theoretical and numerical constraint-handling techniques used with evolutionary algorithms: a survey of the state of the art. Comput Methods Appl Mech Eng 2002;191:1245–87. https://doi.org/https://doi.org/10.1016/S0045-7825(01)00323-1.

[97] Sandgren E. Nonlinear integer and discrete programming in mechanical design optimization. J Mech Des Trans ASME 1990;112:223–9. https://doi.org/10.1115/1.2912596.

[98] Mohapatra S, Mohapatra P. American zebra optimization algorithm for global optimization problems. vol. 13. Nature Publishing Group UK; 2023. https://doi.org/10.1038/s41598-023-31876-2.

[99] Gandomi AH, Yang XS, Alavi AH. Erratum: Cuckoo search algorithm: A metaheuristic approach to solve structural optimization problems (Engineering with Computers DOI:10.1007/s00366-011-0241-y). Eng Comput 2013;29:245. https://doi.org/10.1007/s00366-012-0308-4.

[100] Ray T, Saini P. Engineering design optimization using a swarm with an intelligent information sharing among individuals. Eng Optim 2001;33:735–48. https://doi.org/10.1080/03052150108940941.

[101] Jin N, Rahmat-Samii Y. Advances in particle swarm optimization for antenna designs: Real-number, binary, single-objective and multiobjective implementations. IEEE Trans Antennas Propag 2007;55:556–67. https://doi.org/10.1109/TAP.2007.891552.

[102] Herrera F, Lozano M. Gradual distributed real-coded genetic algorithms. IEEE Trans Evol Comput 2000;4:43–62. https://doi.org/10.1109/4235.843494.

[103] Das S, Suganthan PN. Problem Definitions and Evaluation Criteria for CEC 2011 Competition on Testing Evolutionary Algorithms on Real World Optimization Problems. Electronics 2011:1–42.

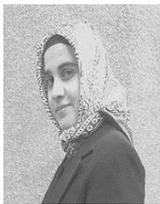
**Hanan K. AbdulKarim** received B.Sc. (Hons.) and M.Sc. degrees in Software Engineering from Salahaddin University- Erbil Iraq, in 2008 and 2014 respectively. She is currently pursuing a Ph.D. in Software Engineering from Salahaddin University, College of Engineering, Erbil, Iraq in swarm intelligence. She worked as a teaching assistant for two years in 2009, then started her M.Sc. study, and worked as an IT manager and instructor in the same university from 2014, where currently studying.

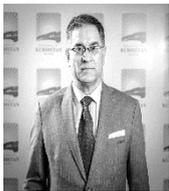
**TARIK A. RASHID** received a Ph.D. degree in computer science and informatics from the College of Engineering, Mathematical and Physical Sciences, University College Dublin (UCD), Ireland in 2006. He joined the University of Kurdistan Hewlêr, in 2017. He is a Principal Fellow for the Higher Education Authority (PFHEA-UK) and a professor in the Department of Computer Science and Engineering at the University of Kurdistan Hewlêr (UKH), Iraq. He is on the prestigious Stanford University list of the World's Top 2% of Scientists for the years 2021, 2022, and 2023. His areas of research cover the fields of Artificial Intelligence, Nature Inspired Algorithms, Swarm Intelligence, Computational Intelligence, Machine Learning, and Data Mining.